\documentclass{article} 
\usepackage{iclr2027_conference,times}

\usepackage{amsmath,amsfonts,bm}

\def\eqref#1{equation~\ref{#1}}

\def\1{\bm{1}}

\DeclareMathAlphabet{\mathsfit}{\encodingdefault}{\sfdefault}{m}{sl}
\SetMathAlphabet{\mathsfit}{bold}{\encodingdefault}{\sfdefault}{bx}{n}

\usepackage{hyperref}
\usepackage{url}
\usepackage{enumitem}
\usepackage{graphicx}
\usepackage{amssymb}
\usepackage{booktabs}
\usepackage[table]{xcolor}
\usepackage{wrapfig}
\definecolor{color4}{RGB}{231, 224, 241}  

\definecolor{headergray}{gray}{0.93}

\usepackage{algorithm}
\usepackage{tabularx}
\usepackage{algorithmicx}
\usepackage{algpseudocode}
\usepackage{xcolor}
\usepackage[most]{tcolorbox}

\definecolor{PromptBlue}{HTML}{4F6D8A}
\definecolor{PromptBlueDark}{HTML}{3E5870}
\definecolor{PromptBlueLight}{HTML}{EEF3F7}

\definecolor{PromptGray}{HTML}{F7F8FA}
\definecolor{PromptLine}{HTML}{D6DEE5}
\definecolor{PromptText}{HTML}{263238}

\definecolor{PromptGreen}{HTML}{5F7F72}
\definecolor{PromptGreenLight}{HTML}{EEF4F1}

\definecolor{PromptOrange}{HTML}{9A7654}
\definecolor{PromptOrangeLight}{HTML}{F8F3EE}

\newcommand{\slot}[1]{%
    \texttt{\textcolor{PromptBlueDark}{\{#1\}}}%
}

\newtcolorbox{promptbox}[1]{
    enhanced,
    breakable,
    colback=PromptGray,
    colframe=PromptLine,
    coltext=PromptText,
    boxrule=0.5pt,
    leftrule=2.2pt,
    colframe=PromptBlue,
    arc=1.2mm,
    left=3mm,
    right=3mm,
    top=2.5mm,
    bottom=2.5mm,
    title=#1,
    fonttitle=\bfseries\small,
    coltitle=white,
    colbacktitle=PromptBlue,
    attach boxed title to top left={
        xshift=2mm,
        yshift=-2mm
    },
    boxed title style={
        colback=PromptBlue,
        colframe=PromptBlue,
        boxrule=0pt,
        arc=1mm,
        left=2mm,
        right=2mm,
        top=0.7mm,
        bottom=0.7mm
    },
    before skip=9pt,
    after skip=9pt
}

\newcommand{\prompttask}[1]{%
    \vspace{1mm}
    \noindent
    \colorbox{PromptBlueLight}{%
        \parbox{\dimexpr\linewidth-2\fboxsep\relax}{%
            \textbf{\textcolor{PromptBlueDark}{Task.}} #1
        }%
    }%
    \vspace{1mm}
}

\newcommand{\promptinput}[1]{%
    \textbf{\textcolor{PromptBlueDark}{#1}}
}

\newcommand{\promptconstraint}[1]{%
    \textbf{\textcolor{PromptOrange}{#1}}
}

\newcommand{\promptoutput}[1]{%
    \vspace{1mm}
    \noindent
    \colorbox{PromptGreenLight}{%
        \parbox{\dimexpr\linewidth-2\fboxsep\relax}{%
            \textbf{\textcolor{PromptGreen}{Output.}} #1
        }%
    }%
}

\title{BIRD: Distilling Decision Boundaries into Rationales for MLLM Adaptation}

\author{
Anglin Liu$^{1}$, Yanlin Wu$^{1}$, Ruichao Chen$^{2}$, Yuting Zhang$^{1}$, 
Qingyuan Zeng$^{1}$, Pengxiang Cai$^{1}$, \\
\textbf{Ziqi Gong}$^{1}$, \textbf{Muchen Li}$^{1}$, \textbf{Jintai Chen}$^{1,2,}$\thanks{
  Corresponding author. Email: \texttt{jintaiCHEN@hkust-gz.edu.cn}.
}  \\
$^{1}$HKUST(GZ),
$^{2}$HKUST
}
\iclrfinalcopy 
\begin{document}

\maketitle
\fancyhead{}
\renewcommand{\headrulewidth}{0pt}

\begin{abstract}
Adapting general-purpose multimodal large language models (MLLMs) to specialized domains requires learning domain-specific decision criteria, which often hinge on subtle visual distinctions between otherwise plausible answers. Rationale augmentation aims to expose such evidence through additional observations or inter-sample comparisons, yet a visually valid cue is not necessarily decision-relevant: it may describe how samples differ without changing the model's relative preference between competing answers. We therefore introduce \textbf{BIRD}, a self-improving \textbf{B}oundary-\textbf{I}nformed \textbf{R}ationale \textbf{D}istillation framework that uses model-specific confusions to locate unresolved local decision boundaries and distills the evidence that resolves these confusions into rationales. For each sample, BIRD retrieves candidate neighbors from the target MLLM's own representation space and selects the most confusable one according to its answer preferences. It then generates answer-blind candidate evidence from their visual differences and functionally verifies which evidence most effectively strengthens the model's preference for the correct answer while avoiding inappropriate transfer across the pair. The verified evidence is then distilled into a single-sample rationale for standard supervised fine-tuning. Experiments on medical and chart VQA show that BIRD outperforms competing rationale-augmentation methods across two target MLLMs, while further analyses demonstrate clearer separation of confusable answers and stronger gains from model-matched supervision.
\end{abstract}

\section{Introduction}
\label{sec:introduction}

General-purpose multimodal large language models (MLLMs) possess broad visual-language capabilities~\citep{li2023blip, liu2023visual, dai2023instructblip}, but specialized domains often hinge on distinctions that are subtle, domain-dependent, and decisive for the final prediction~\citep{li2023llava, tu2024towards, masry2024chartinstruct}. In medical image analysis, for example, visually similar findings may indicate different diagnoses because of subtle differences in morphology or spatial distribution; in scientific chart understanding, a small trend reversal or relative change may alter the correct conclusion. At their core, specialized domains differ in the criteria that determine which visual distinctions are decision-defining. Rationales provide a natural way to express such criteria by making decision-relevant visual evidence explicit, thereby offering richer supervision than answer labels alone~\citep{park2018multimodal, kayser2021vil, sammani2022nlx}.

\begin{figure}[t]
    \centering
    \includegraphics[width=0.85\linewidth]{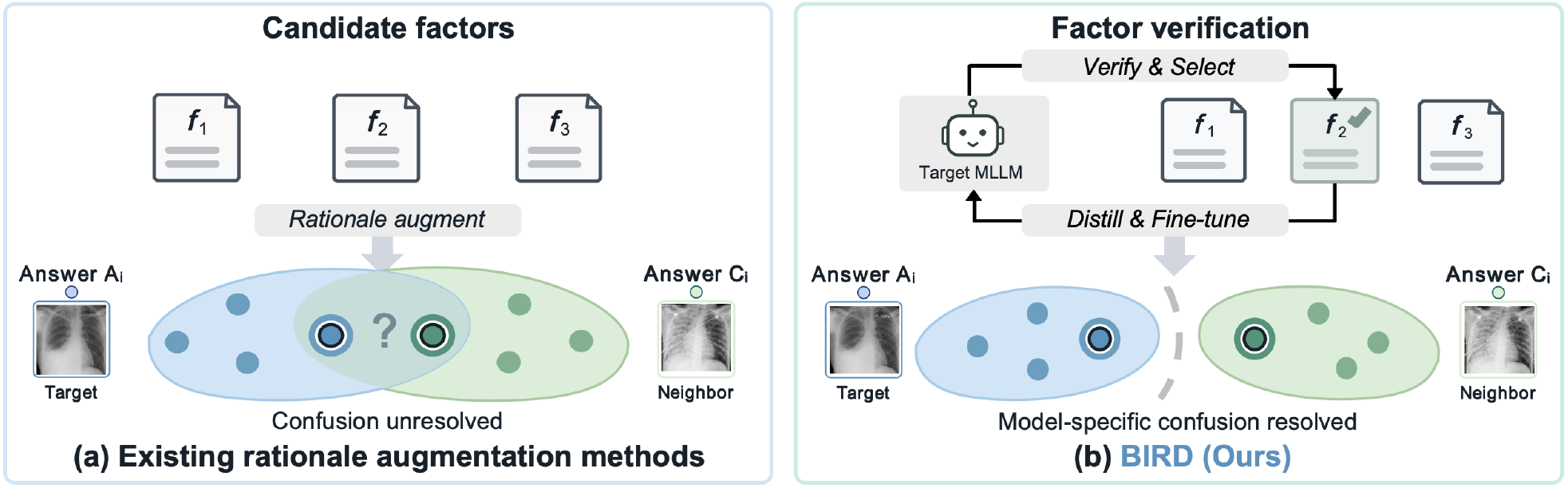}
    \caption{Existing rationale augmentation methods leave confusable answers unresolved. BIRD verifies candidate factors with the target MLLM and distills the selected factor to resolve model-specific confusion.}
    \label{fig:fig1}
    \vspace{-0.4cm}
\end{figure}

Recent methods have therefore sought to improve domain-specialized MLLMs by augmenting rationale supervision~\citep{carbune-etal-2024-chart, zhu-etal-2024-efficient, zhang-etal-2025-improve, WanYua_V2TCoT_MICCAI2025}, where additional evidence is derived either from individual samples or from relations between samples (Figure~\ref{fig:fig1}(a)). The latter is especially useful when visually similar samples lead to different answers, as their comparison can expose subtle distinctions that are difficult to identify from either sample alone, motivating recent inter-sample approaches~\citep{zou2025alignment, xiong2026retrieving}. However, a visible difference is not necessarily relevant to the model's decision---a cue may accurately distinguish two samples while leaving their relative answer preferences unchanged. Such a cue adds descriptive detail to the rationale without providing evidence that resolves the model's ``{confusion}''. Effective rationale augmentation should therefore identify the evidence that changes the competition between plausible answers, increasing the model's preference for the correct answer while suppressing its confusable alternatives. Such ``confusion'' reveals where the model's local decision boundaries remain unresolved.

Equally importantly, these unresolved decision boundaries are model-specific. Different MLLMs organize samples into different visual neighborhoods, confuse different alternatives, and rely on different cues to resolve those confusions. A pair that lies near an unresolved boundary for one model may be readily distinguished by another; even for the same pair, the evidence needed to resolve the confusion may differ across models.
Existing model-agnostic rationale augmentation methods~\citep{chen2024your, xiong2025hs, wu2025towards} cannot consistently target the specific weaknesses of a given MLLM.
This motivates a self-improving formulation in which the target MLLM identifies its own confusions, locates the corresponding unresolved decision boundaries, and converts the evidence needed to resolve them into rationale supervision (Figure~\ref{fig:fig1}(b)).

Building on this perspective, we propose \textbf{BIRD}, a self-improving \textbf{B}oundary-\textbf{I}nformed \textbf{R}ationale \textbf{D}istillation framework that converts model-specific confusions into rationale supervision. For each training sample, BIRD retrieves semantically compatible samples with different answers from the target model's visual space and uses its answer preferences to identify the most confusable neighbor. The model then performs an answer-blind comparison to generate question-relevant visual differences as candidate evidence. Rather than treating these differences as equally useful, BIRD evaluates how each candidate changes the model's preference across the pair and selects the one that most effectively resolves the confusion while preserving the distinction between the two sides. The verified evidence is then distilled into a single-sample rationale, allowing the target MLLM to turn its own unresolved decision boundaries into supervision for adaptation.
Experiments on medical and chart VQA show that BIRD achieves the strongest average performance across two target MLLMs, outperforming competing rationale-augmentation methods on in-domain benchmarks. Further analyses reveal clearer separation between confusable answers, boundary-specific effects of the selected factors, and stronger gains from model-matched supervision.

Our contributions are threefold:


\begin{itemize}[leftmargin=*]

\item \textit{Conceptualizing rationale augmentation as model-specific boundary resolution.}
We recast rationale augmentation from enriching sample descriptions to resolving the target MLLM's unresolved decision boundaries, turning rationale construction into model-conditioned evidence selection based on whether evidence resolves the model's own confusions.

\item \textit{Self-improving distillation of model-specific boundary evidence.}
We introduce BIRD, which discovers model-specific confusable neighbors, evaluates answer-blind candidate evidence by own-side gain and cross-boundary transfer, and distills only verified evidence into single-sample rationales for standard supervised fine-tuning.

\item \textit{Demonstrating the advantage of model-matched boundary supervision.}
Across two target MLLMs and medical and chart VQA, BIRD achieves the best overall average on most model--benchmark combinations, with analyses showing sharper separation of confusable answers, boundary-specific evidence effects, and stronger gains from model-matched supervision.

\end{itemize}

\section{Related Work}
\label{sec:related}

\textbf{Self-Improving Multimodal Reasoning.}
Building on STaR's perspective~\citep{zelikman2022star}, recent work has explored improving multimodal reasoning by turning model-generated
solutions into supervision.
$R^3V$ iteratively bootstraps positive and negative multimodal reasoning traces
and learns to refine or select rationales through self-reflection~\citep{cheng2025vision}.
M-STAR systematically studies self-evolving multimodal training through the
choice of training objective, reward model, and prompt variation, and introduces
adaptive balancing to alleviate performance saturation~\citep{liu2024diving}.
VC-STaR exploits visually similar VQA pairs
to help VLMs identify visual discrepancies,
thereby converting the model's contrastive ability into improved reasoning
supervision~\citep{pan2026through}.
Concurrent work further improves self-training by explicitly verifying perceptual
grounding~\citep{sharma2026improving}.
Despite these advances, existing self-improving methods mainly bootstrap better reasoning from model-generated trajectories or feedback, but do not explicitly target the model's own unresolved decision boundaries.

\textbf{Inter-Sample Supervision.}
Building on complementary and counterfactual VQA, inter-sample supervision has
been used to explicitly model the relationship among factual, original, and
counterfactual samples~\citep{liang2020learning}. Hard-negative mining further
exploits confusable instances to improve cross-modal discrimination in medical
VQA, while counterfactual image substitution has been
used to measure and strengthen visual reliance~\citep{zafar2026medical}. More recently, counterfactual-style retrieval has
been used to select causally informative demonstrations for visual in-context
learning~\citep{xiong2026retrieving}. However, these methods don't test whether a difference actually changes the target model’s preference between competing answers.

\section{Pilot Study: Probing Model-Specific Boundaries}
\label{sec:pilot}

Before introducing BIRD, we conduct a lightweight pilot study with Qwen3.5-9B~\citep{qwen35blog} and InternVL3-8B~\citep{zhu2025internvl3} to examine two questions: whether local confusion is specific to the target MLLM, and whether visual differences between similar samples necessarily provide useful boundary evidence.

\begin{figure}[h]
    \centering
    \includegraphics[width=1\linewidth]{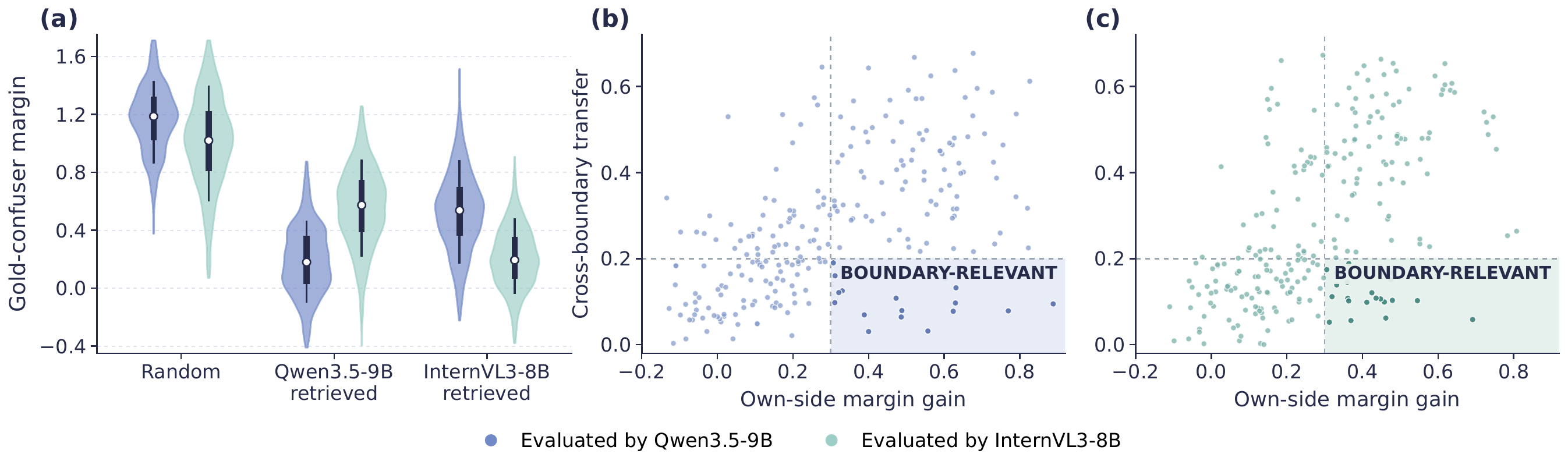}
    \vskip -1 em
    \caption{Pilot study of model-specific boundaries.
(a) Model-retrieved neighbors yield lower gold--confuser margins than random pairs, with each model most confused by neighbors retrieved from its own representation space.
(b,c) Candidate factors exhibit heterogeneous decision effects, only a small subset falls in the shaded boundary-relevant region.}
    \label{fig:fig2}
\end{figure}

\textbf{Hard near neighbors are model-specific.}
For each anchor sample $i$, we retrieve its nearest different-answer neighbor using the target MLLM's own visual representation, and compare it with a random different-answer sample from the same compatible pool. Pair difficulty is measured by the gold--confuser margin:
\begin{equation}
m_M(i,j)
=
\ell_M(A_i \mid I_i,Q_i)
-
\ell_M(A_j \mid I_i,Q_i),
\end{equation}
where $\ell_M$ is the length-normalized answer log-likelihood, and a smaller margin indicates stronger confusion.
We repeat the retrieval independently with two MLLMs and evaluate all pairs under both models. As shown in Figure~\ref{fig:fig2}(a), nearest different-answer neighbors are consistently harder than random compatible pairs. Moreover, each model is most strongly confused by neighbors retrieved from its own representation space, suggesting that unresolved local boundaries are model-specific rather than fixed properties of the dataset.

\textbf{Pairwise differences are not necessarily boundary evidence.}
For each hard pair, we hide the answers and ask the corresponding target MLLM to propose question-relevant visual differences. For each candidate factor $e$, we measure its \emph{own-side gain}, i.e., the increase in the correct-versus-confuser margin on the image where it was observed, and its \emph{cross-boundary transfer}, i.e., whether the same evidence also pushes the opposite-side image toward the anchor's answer:
\begin{align}
\Delta_{\mathrm{own}}
&=
m_M(I_i,e;A_i,A_j,Q_i)
-
m_M(I_i,\varnothing;A_i,A_j,Q_i),\\
\Delta_{\mathrm{cross}}
&=
\Big[
m_M(I_j,e;A_i,A_j,Q_i)
-
m_M(I_j,\varnothing;A_i,A_j,Q_i)
\Big]_+ .
\end{align}
We conduct this analysis separately for the hard pairs discovered by each target MLLM. Figures~\ref{fig:fig2}(b) and (c) show the candidate evidence distributions for Qwen3.5-9B and InternVL3-8B, respectively. Both models exhibit substantial, yet model-specific, heterogeneity: many visible differences have little effect on the decision, while others increase the desired margin but transfer similarly across the paired sample. For each model, only a subset exhibits the desired behavior of high own-side gain and low cross-boundary transfer.

\textbf{Implication.}
These observations motivate two design principles:
The target MLLM should discover its own confusable neighbors, and pairwise differences should be treated only as candidate evidence until their boundary relevance is functionally verified.
BIRD follows these principles by discovering model-specific confusers, verifying discriminative factors, and distilling the validated evidence into sample-specific rationales.


\section{BIRD: Boundary-Informed Rationale Distillation}
\label{sec:method}

\subsection{Overview}

\begin{figure}[t]
    \centering
    \includegraphics[width=1\linewidth]{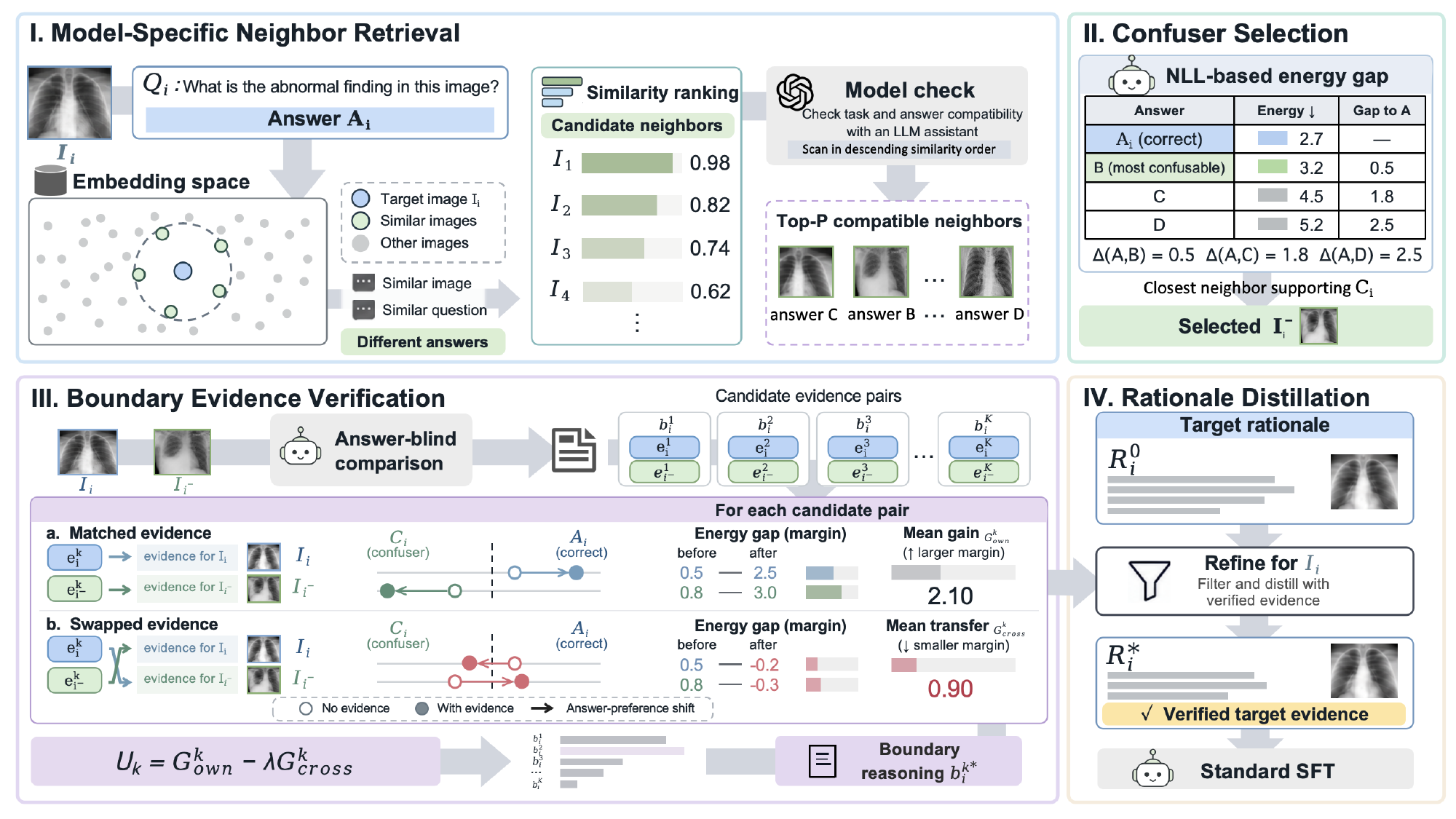}
    \caption{Overview of BIRD. The target MLLM retrieves model-specific neighbors, selects its most confusable alternative, verifies candidate boundary evidence with energy-based gain and cross-boundary transfer, and distills the validated evidence into a refined rationale for standard SFT.}
    \label{fig:fig3}
\end{figure}

The pipeline of BIRD is shown in Figure~\ref{fig:fig3}. Given a domain-specific training set
\begin{equation}
    \mathcal{D}_0=\{(I_i,Q_i,R_i^0,A_i)\}_{i=1}^{N}
\end{equation}
and a target MLLM $M_\theta$, BIRD improves the training rationales while
keeping the images, questions, answers, and sample size unchanged.
For each sample, BIRD retrieves model-specific neighbors, identifies the
alternative most confusable to the target model, proposes and verifies
boundary evidence, and distills the verified evidence into the rationale:
\begin{equation}
    \mathcal{D}^{*}=\{(I_i,Q_i,R_i^{*},A_i)\}_{i=1}^{N}.
\end{equation}
The target MLLM’s visual representations define the candidate neighborhoods, its answer preferences identify confusable alternatives and score candidate factors, and the selected factor is distilled into supervision for improving the same model.

\subsection{Model-Specific Neighbor Retrieval}

Our pilot study shows difficult local neighbors are model-specific.
We therefore retrieve them using the target MLLM's own visual representation.
Let $h^{\mathrm{vis}}_{i,t}$ denote the projected visual tokens of $I_i$:
\begin{equation}
    v_i =
    \operatorname{Norm}
    \left(
    \frac{1}{T_i}\sum_{t=1}^{T_i} h^{\mathrm{vis}}_{i,t}
    \right).
\end{equation}

For each sample, we first restrict retrieval to semantically relevant questions
using question templates when available or question similarity otherwise, and
then retrieve the top-$P$ samples by $\cos(v_i,v_j)$, while using GPT-5 to check whether these samples are performing the same task.
Thus, the target MLLM itself defines the local neighborhoods in which BIRD searches for unresolved confusions.

We further remove QA pairs that cannot form a meaningful comparison with the
target question. When question formulations differ, a neighbor is retained
only if its known QA state can be unambiguously expressed as an alternative
answer to the target question; otherwise it is discarded.

\subsection{Confuser Selection}

Among the retrieved neighbors, we follow the energy-based view~\citep{liu2026med} and identify the alternative that is most confusable to the target model by quantifying the model’s preference,
\begin{equation}
    E_\theta(a\mid I,Q)
    =
    -\frac{1}{|a|}
    \sum_{t=1}^{|a|}
    \log p_\theta(a_t\mid I,Q,a_{<t}),
    \label{eq:energy}
\end{equation}
where lower energy indicates stronger model preference.

Let $\mathcal{A}_i^{-}$ be the alternative answers provided by the retrieved
neighbors in the answer space of $Q_i$. We select
\begin{equation}
    C_i
    =
    \arg\min_{c\in\mathcal{A}_i^{-}}
    E_\theta(c\mid I_i,Q_i).
    \label{eq:confuser}
\end{equation}
Hence, the confuser is determined directly by the target MLLM's current energy
landscape. Among neighbors corresponding to $C_i$, we choose the visually
closest one as the witness image $I_i^{-}$.
The resulting pair $(I_i,I_i^{-})$ probes a local boundary that is difficult
for the current model.

\subsection{Boundary Evidence Verification}

A confusable pair only suggests where a boundary may lie, and not every visible
difference is decision-relevant. We therefore hide both answers and ask the target MLLM to propose $K$ question-relevant visual factors. Finally, we get:
\begin{equation}
    b_i^k=(e_i^k,e_{i^-}^k), \qquad k=1,\ldots,K,
\end{equation}
where $e_i^k$ and $e_{i^-}^k$ describe corresponding observations on the two
images. $K$ is set to 3 by default. Candidates that explicitly reveal an answer are discarded.

We verify each factor through its effect on the target model's energy gap:
\begin{equation}
    g_\theta(I,e;A,C,Q)
    =
    E_\theta(C\mid I,Q,e)
    -
    E_\theta(A\mid I,Q,e),
    \label{eq:energy_gap}
\end{equation}
where a larger value indicates stronger preference for $A$ over $C$.

Its own-side gain is
\begin{align}
    \Delta_t^k &=
    g_\theta(I_i,e_i^k;A_i,C_i,Q_i)
    -
    g_\theta(I_i,\varnothing;A_i,C_i,Q_i),\\
    \Delta_n^k &=
    g_\theta(I_i^{-},e_{i^-}^k;C_i,A_i,Q_i)
    -
    g_\theta(I_i^{-},\varnothing;C_i,A_i,Q_i),
\end{align}
with
\begin{equation}
    G_{\mathrm{own}}^k
    =
    \frac{\Delta_t^k+\Delta_n^k}{2}.
\end{equation}

We then swap the evidence across the pair to measure whether it transfers to
the wrong side:
\begin{align}
    L_{t\rightarrow n}^{k}
    &=
    \Big[
    g_\theta(I_i^{-},e_i^k;A_i,C_i,Q_i)
    -
    g_\theta(I_i^{-},\varnothing;A_i,C_i,Q_i)
    \Big]_+,\\
    L_{n\rightarrow t}^{k}
    &=
    \Big[
    g_\theta(I_i,e_{i^-}^k;C_i,A_i,Q_i)
    -
    g_\theta(I_i,\varnothing;C_i,A_i,Q_i)
    \Big]_+,
\end{align}
and
\begin{equation}
    G_{\mathrm{cross}}^k
    =
    \frac{L_{t\rightarrow n}^{k}+L_{n\rightarrow t}^{k}}{2}.
\end{equation}

We define the \emph{boundary utility} as
\begin{equation}
    U_k
    =
    G_{\mathrm{own}}^k
    -
    \lambda G_{\mathrm{cross}}^k.
    \label{eq:boundary_utility}
\end{equation}
where $\lambda$ is the cross-boundary penalty. A high-utility factor strengthens the appropriate decision on its own side
while avoiding the same effect across the boundary. We select
$k^*=\arg\max_k U_k$.
Because both proposal and verification are performed by the target MLLM,
the selected evidence specifically addresses its current decision weakness.

\subsection{Rationale Distillation}

Finally, BIRD converts the verified evidence into supervision for improving the
same target model. If the best factor has sufficient boundary utility, we refine
the original rationale as
\begin{equation}
    R_i^*
    =
    \operatorname{Refine}
    \left(
        R_i^0;
        I_i,Q_i,A_i,C_i,
        e_i^{k^*},e_{i^-}^{k^*}
    \right).
\end{equation}
The refined rationale preserves valid information in $R_i^0$, incorporates the
verified target-side evidence, and introduces no unsupported visual observations.
Neighbor evidence is used only as contrastive context.

If no factor reaches the utility threshold $\tau_U$, we retain $R_i^*=R_i^0$.
We then perform standard rationale-supervised fine-tuning on $\mathcal{D}^*$.
No additional contrastive objective, preference optimization, or reinforcement
learning is required, and no retrieval or paired image is used at training or inference.

\section{Experiments and Analysis}

\subsection{Experimental Setup}
\label{sec:exp_setup}

\textbf{Models and benchmarks.}
We evaluate BIRD with two target MLLMs, Qwen3.5-9B and InternVL3-8B,
under two independent domain-adaptation settings.
For chart and plot reasoning, we adapt each model on the ChartQA-X~\citep{hegde2025chartqaxgeneratingexplanationsvisual} training set
and evaluate on ChartQA~\citep{masry-etal-2022-chartqa}, ChartBench~\citep{xu2023chartbench}, and ChartQAPro~\citep{masry2025chartqapro}, which cover diverse
chart-understanding and numerical reasoning tasks.
For medical reasoning, we use OpenMedReason~\citep{baghbanzadeh2026openmedreason} for adaptation and evaluate on SLAKE~\citep{liu2021slake},
PathVQA~\citep{he2021towards}, and MedXpertQA~\citep{zuo2025medxpertqa}, spanning general medical VQA, pathology-focused reasoning, and challenging medical visual question answering.

\textbf{Baselines.}
We compare against the untuned base model and Original SFT, which directly fine-tunes on the original domain data without rationale enhancement.
We further include representative rationale-enhancement methods: Reflective Instruction Tuning (Reflective IT)~\citep{zhang2024reflective}, which augments rationale supervision without self-improvement; STaR, R$^3$V, and Self-Rationale Calibration (SRC)~\citep{wu2025towards}, which improve supervision using the target model's own
reasoning or responses; and VC-STaR, which additionally exploits inter-sample visual contrast. These methods provide comparisons across both self-improving and non-self-improving settings, with and without inter-sample supervision.

\textbf{Training protocol.}
All methods use the same source data and target MLLM within each model--domain
setting. We train for 10 epochs on 8 NVIDIA A800 GPUs with AdamW
($\mathrm{lr}=2\times10^{-5}$, global batch size $=128$), using a cosine
learning-rate schedule with a 3\% warmup ratio. For BIRD, we set the
cross-boundary penalty to $\lambda=2.0$ and the utility threshold to
$\tau_U=0.2$, so that only candidates with boundary utility above 0.2 are distilled into the training rationales. $K$ and $P$ are both set to 3 by default.
Full details are provided in the Appendix.

\subsection{Overall Domain Adaptation Performance}
\label{sec:main_results}

Table~\ref{tab:main_results} reports the adaptation performance across the two target MLLMs and domains. BIRD achieves the highest overall averages, improving the base models by 4.97 and 5.05 points for both models, respectively. It further surpasses the strongest competing rationale-augmentation method, VC-STaR. These gains suggest that verified boundary-specific evidence provides more effective adaptation supervision.

\begin{table*}[t]
\centering
\caption{
Main results across two target MLLMs and two domain-adaptation settings.
Higher is better. ``Self'' indicates self-improvement, while ``Inter'' denotes inter-sample supervision. Arrows indicate absolute changes relative to the corresponding base model without SFT.
}
\label{tab:main_results}
\scriptsize
\setlength{\tabcolsep}{2.8pt}
\renewcommand{\arraystretch}{1.04}
\begin{tabular}{l|c|c|ccc|ccc|c}
\toprule
& & &
\multicolumn{3}{c|}{\textbf{Chart / Plot VQA}} &
\multicolumn{3}{c|}{\textbf{Medical VQA}} \\
\cmidrule(lr){4-6}\cmidrule(lr){7-9}\cmidrule(lr){10-10}
\textbf{Method}
& \textbf{Self}
& \textbf{Inter}
& \textbf{ChartQA}
& \textbf{ChartBench}
& \textbf{ChartQAPro}
& \textbf{SLAKE}
& \textbf{PathVQA}
& \textbf{MedXpertQA}
& \textbf{Avg.}\\
\midrule

\rowcolor{headergray} \multicolumn{10}{@{}l}{\textit{Qwen3.5-9B}} \\
Base (no SFT)
& & &
85.67 & 67.42 & 41.35 & 73.23 & 47.12 & 27.18 & 57.00 \\

Original SFT
& & &
88.36 {\tiny\textcolor{blue}{$\uparrow$\,2.69}} &
68.20 {\tiny\textcolor{blue}{$\uparrow$\,0.78}} &
49.02 {\tiny\textcolor{blue}{$\uparrow$\,7.67}} &
75.49 {\tiny\textcolor{blue}{$\uparrow$\,2.26}} &
49.53 {\tiny\textcolor{blue}{$\uparrow$\,2.41}} &
27.84 {\tiny\textcolor{blue}{$\uparrow$\,0.66}} &
59.74 {\tiny\textcolor{blue}{$\uparrow$\,2.74}} \\

Reflective IT
& & &
88.48 {\tiny\textcolor{blue}{$\uparrow$\,2.81}} &
68.17 {\tiny\textcolor{blue}{$\uparrow$\,0.75}} &
50.33 {\tiny\textcolor{blue}{$\uparrow$\,8.98}} &
75.81 {\tiny\textcolor{blue}{$\uparrow$\,2.58}} &
49.35 {\tiny\textcolor{blue}{$\uparrow$\,2.23}} &
27.92 {\tiny\textcolor{blue}{$\uparrow$\,0.74}} &
60.01 {\tiny\textcolor{blue}{$\uparrow$\,3.01}} \\ \hline

STaR
& \checkmark & &
87.64 {\tiny\textcolor{blue}{$\uparrow$\,1.97}} &
67.99 {\tiny\textcolor{blue}{$\uparrow$\,0.57}} &
49.26 {\tiny\textcolor{blue}{$\uparrow$\,7.91}} &
75.19 {\tiny\textcolor{blue}{$\uparrow$\,1.96}} &
49.40 {\tiny\textcolor{blue}{$\uparrow$\,2.28}} &
27.83 {\tiny\textcolor{blue}{$\uparrow$\,0.65}} &
59.55 {\tiny\textcolor{blue}{$\uparrow$\,2.55}} \\

R$^3$V
& \checkmark & &
88.26 {\tiny\textcolor{blue}{$\uparrow$\,2.59}} &
68.12 {\tiny\textcolor{blue}{$\uparrow$\,0.70}} &
47.39 {\tiny\textcolor{blue}{$\uparrow$\,6.04}} &
75.25 {\tiny\textcolor{blue}{$\uparrow$\,2.02}} &
49.78 {\tiny\textcolor{blue}{$\uparrow$\,2.66}} &
26.98 {\tiny\textcolor{red}{$\downarrow$\,0.20}} &
59.30 {\tiny\textcolor{blue}{$\uparrow$\,2.30}} \\

SRC
& \checkmark & &
83.62 {\tiny\textcolor{red}{$\downarrow$\,2.05}} &
67.15 {\tiny\textcolor{red}{$\downarrow$\,0.27}} &
46.67 {\tiny\textcolor{blue}{$\uparrow$\,5.32}} &
74.91 {\tiny\textcolor{blue}{$\uparrow$\,1.68}} &
48.66 {\tiny\textcolor{blue}{$\uparrow$\,1.54}} &
26.65 {\tiny\textcolor{red}{$\downarrow$\,0.53}} &
57.94 {\tiny\textcolor{blue}{$\uparrow$\,0.94}} \\ \hline

VC-STaR
& \checkmark & \checkmark &
89.45 {\tiny\textcolor{blue}{$\uparrow$\,3.78}} &
69.20 {\tiny\textcolor{blue}{$\uparrow$\,1.78}} &
51.38 {\tiny\textcolor{blue}{$\uparrow$\,10.03}} &
75.96 {\tiny\textcolor{blue}{$\uparrow$\,2.73}} &
50.29 {\tiny\textcolor{blue}{$\uparrow$\,3.17}} &
28.10 {\tiny\textcolor{blue}{$\uparrow$\,0.92}} &
60.73 {\tiny\textcolor{blue}{$\uparrow$\,3.73}} \\ \hline

\rowcolor{color4}\textbf{BIRD}
& \checkmark & \checkmark &
89.84 {\tiny\textcolor{green!60!black}{$\uparrow$\,4.17}} &
71.55 {\tiny\textcolor{green!60!black}{$\uparrow$\,4.13}} &
52.96 {\tiny\textcolor{green!60!black}{$\uparrow$\,11.61}} &
77.92 {\tiny\textcolor{green!60!black}{$\uparrow$\,4.69}} &
51.49 {\tiny\textcolor{green!60!black}{$\uparrow$\,4.37}} &
28.06 {\tiny\textcolor{green!60!black}{$\uparrow$\,0.88}} &
61.97 {\tiny\textcolor{green!60!black}{$\uparrow$\,4.97}} \\

\midrule
\rowcolor{headergray} \multicolumn{10}{@{}l}{\textit{InternVL3-8B}} \\

Base (no SFT)
& & &
82.72 & 65.07 & 37.78 & 72.83 & 48.64 & 22.39 & 54.91 \\

Original SFT
& & &
85.31 {\tiny\textcolor{blue}{$\uparrow$\,2.59}} &
66.55 {\tiny\textcolor{blue}{$\uparrow$\,1.48}} &
47.26 {\tiny\textcolor{blue}{$\uparrow$\,9.48}} &
74.13 {\tiny\textcolor{blue}{$\uparrow$\,1.30}} &
50.08 {\tiny\textcolor{blue}{$\uparrow$\,1.44}} &
23.43 {\tiny\textcolor{blue}{$\uparrow$\,1.04}} &
57.79 {\tiny\textcolor{blue}{$\uparrow$\,2.88}} \\

Reflective IT
& & &
85.44 {\tiny\textcolor{blue}{$\uparrow$\,2.72}} &
66.71 {\tiny\textcolor{blue}{$\uparrow$\,1.64}} &
48.11 {\tiny\textcolor{blue}{$\uparrow$\,10.33}} &
74.56 {\tiny\textcolor{blue}{$\uparrow$\,1.73}} &
50.21 {\tiny\textcolor{blue}{$\uparrow$\,1.57}} &
23.51 {\tiny\textcolor{blue}{$\uparrow$\,1.12}} &
58.09 {\tiny\textcolor{blue}{$\uparrow$\,3.18}} \\ \hline

STaR
& \checkmark & &
84.92 {\tiny\textcolor{blue}{$\uparrow$\,2.20}} &
66.31 {\tiny\textcolor{blue}{$\uparrow$\,1.24}} &
47.58 {\tiny\textcolor{blue}{$\uparrow$\,9.80}} &
74.02 {\tiny\textcolor{blue}{$\uparrow$\,1.19}} &
50.03 {\tiny\textcolor{blue}{$\uparrow$\,1.39}} &
23.36 {\tiny\textcolor{blue}{$\uparrow$\,0.97}} &
57.70 {\tiny\textcolor{blue}{$\uparrow$\,2.79}} \\

R$^3$V
& \checkmark & &
85.18 {\tiny\textcolor{blue}{$\uparrow$\,2.46}} &
66.44 {\tiny\textcolor{blue}{$\uparrow$\,1.37}} &
45.91 {\tiny\textcolor{blue}{$\uparrow$\,8.13}} &
74.11 {\tiny\textcolor{blue}{$\uparrow$\,1.28}} &
50.34 {\tiny\textcolor{blue}{$\uparrow$\,1.70}} &
22.81 {\tiny\textcolor{blue}{$\uparrow$\,0.42}} &
57.47 {\tiny\textcolor{blue}{$\uparrow$\,2.56}} \\

SRC
& \checkmark & &
81.47 {\tiny\textcolor{red}{$\downarrow$\,1.25}} &
65.12 {\tiny\textcolor{blue}{$\uparrow$\,0.05}} &
44.73 {\tiny\textcolor{blue}{$\uparrow$\,6.95}} &
73.62 {\tiny\textcolor{blue}{$\uparrow$\,0.79}} &
49.21 {\tiny\textcolor{blue}{$\uparrow$\,0.57}} &
22.46 {\tiny\textcolor{blue}{$\uparrow$\,0.07}} &
56.10 {\tiny\textcolor{blue}{$\uparrow$\,1.19}} \\ \hline

VC-STaR
& \checkmark & \checkmark &
86.38 {\tiny\textcolor{blue}{$\uparrow$\,3.66}} &
67.53 {\tiny\textcolor{blue}{$\uparrow$\,2.46}} &
49.61 {\tiny\textcolor{blue}{$\uparrow$\,11.83}} &
74.81 {\tiny\textcolor{blue}{$\uparrow$\,1.98}} &
50.86 {\tiny\textcolor{blue}{$\uparrow$\,2.22}} &
23.74 {\tiny\textcolor{blue}{$\uparrow$\,1.35}} &
58.82 {\tiny\textcolor{blue}{$\uparrow$\,3.91}} \\ \hline

\rowcolor{color4}\textbf{BIRD}
& \checkmark & \checkmark &
87.06 {\tiny\textcolor{green!60!black}{$\uparrow$\,4.34}} &
69.42 {\tiny\textcolor{green!60!black}{$\uparrow$\,4.35}} &
51.28 {\tiny\textcolor{green!60!black}{$\uparrow$\,13.50}} &
76.31 {\tiny\textcolor{green!60!black}{$\uparrow$\,3.48}} &
51.97 {\tiny\textcolor{green!60!black}{$\uparrow$\,3.33}} &
23.69 {\tiny\textcolor{green!60!black}{$\uparrow$\,1.30}} &
59.96 {\tiny\textcolor{green!60!black}{$\uparrow$\,5.05}} \\

\bottomrule
\end{tabular}
\vspace{-0.2cm}
\end{table*}

\subsection{Decision Boundary Sharpening}
\label{sec:boundary_sharpening}

\begin{figure}[h]
    \centering
    \includegraphics[width=1\linewidth]{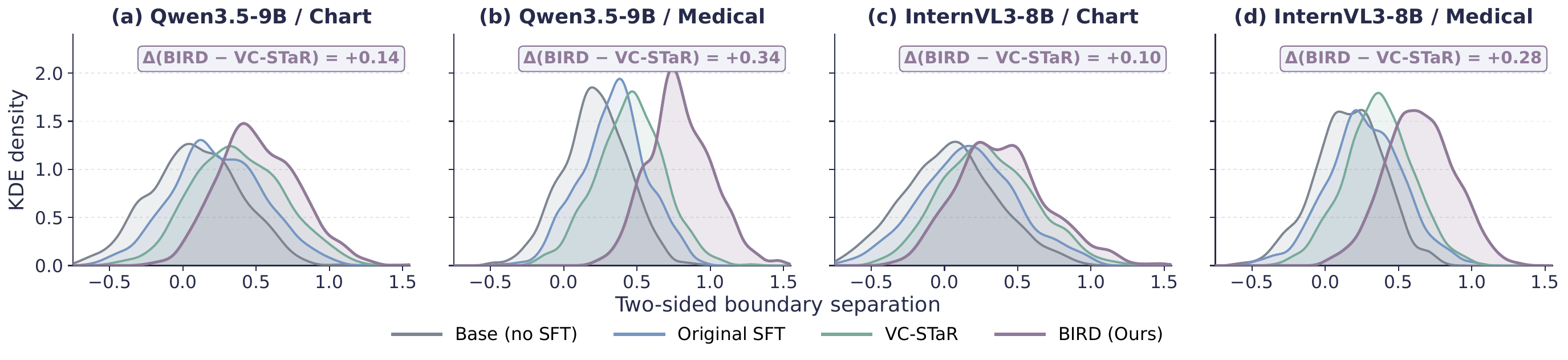}
    \vspace{-0.2cm}
    \caption{Held-out hard-pair distributions of two-sided boundary separation for base, Original SFT, VC-STaR, and BIRD. A consistent rightward shift for BIRD indicates that it more effectively separates the gold answer from the model's original confuser.}
    \label{fig:fig4}
\end{figure}

We further examine whether BIRD sharpens the decision boundaries targeted during adaptation. For each target MLLM, we construct fixed hard pairs from the held-out benchmark splits using Kimi K3~\citep{team2026kimi}.
For each anchor, we retrieve question-compatible samples within the same benchmark, identify its lowest-energy alternative as the confuser, and select the closest valid neighbor supporting that alternative.
We sample 50 hard pairs per benchmark, yielding 150 pairs for each model--domain setting, and use the same pairs to evaluate all methods.

For a pair $(I_i,I_i^-)$ with competing answers $(A_i,C_i)$, we measure two-sided boundary separation as:
\begin{equation}
\begin{split}
S_i=\frac{1}{2}\big[
&E(C_i\mid I_i,Q_i)-E(A_i\mid I_i,Q_i)\\
+&E(A_i\mid I_i^-,Q_i)-E(C_i\mid I_i^-,Q_i)
\big].
\end{split}
\label{eq:boundary_gap}
\end{equation}
Larger $S_i$ indicates a clearer boundary.
Figure~\ref{fig:fig4} compares the distributions of $S_i$ for the base model, Original SFT, the strongest prior baseline, and BIRD across both models and domains.
A consistent rightward shift for BIRD indicates that it more effectively separates the gold answer from the model's original confuser.

\subsection{Model-Specific Self-Improvement}
\label{sec:model_specific}

BIRD is self-improving in that each target MLLM discovers and resolves its own decision ambiguities. We examine whether the resulting supervision is indeed model-specific from both the evidence and downstream perspectives.

\begin{wrapfigure}[15]{r}{0.5\columnwidth}
    \centering
    \includegraphics[width=\linewidth]{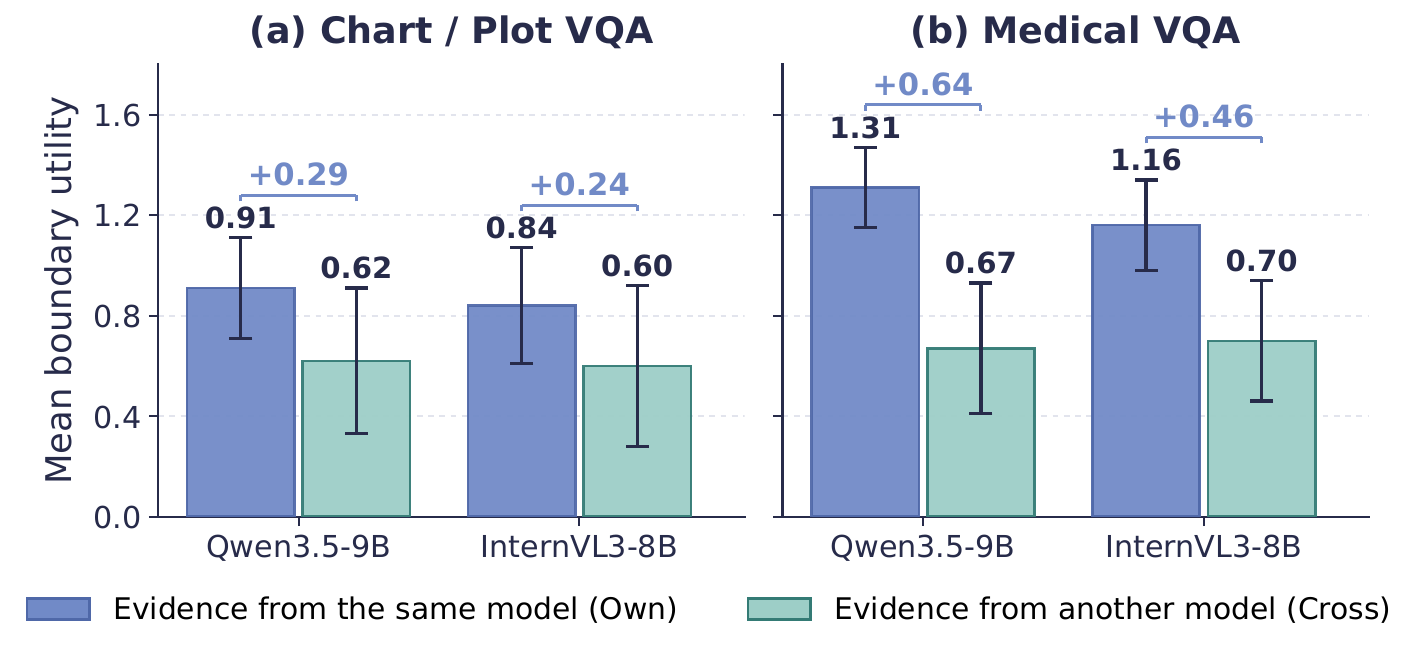}
    \vspace{-0.35cm}
    \caption{Cross-model boundary utility. Evidence discovered by each target MLLM achieves higher utility when evaluated by the same model, demonstrating model-specific boundary supervision.}
    \label{fig:fig5}
\end{wrapfigure}

\textbf{Cross-model evidence utility.}
As shown in Figure~\ref{fig:fig5}, on the same held-out anchors, Qwen3.5-9B and InternVL3-8B independently run BIRD to discover their boundary pairs and selected evidence.
For anchors where both models obtain valid evidence, we re-evaluate the factor discovered by each model using both target MLLMs, while keeping its discovered pair fixed. We then compare the mean boundary utility across evidence sources. If the supervision is model-specific, evidence discovered by a model should yield higher utility when evaluated by that same model.

\textbf{Cross-model supervision transfer.}
We further test whether this specificity translates into downstream adaptation.
As shown in Table~\ref{tab:cross_model_transfer}, for each domain, we use the shared training datasets. Each target MLLM is fine-tuned using rationales produced either by itself or by the other model.
Better performance with self-generated BIRD supervision would provide direct evidence that BIRD converts model-specific weaknesses into supervision that is particularly useful for improving the model itself.

\begin{table*}[h]
\centering
\vspace{-0.2cm}
\caption{
Cross-model transfer of BIRD supervision.
``Own'' denotes rationales constructed by the target MLLM itself, while
``Cross'' uses rationales constructed by the other MLLM.
}
\label{tab:cross_model_transfer}
\scriptsize
\setlength{\tabcolsep}{3.2pt}
\renewcommand{\arraystretch}{1.05}
\begin{tabular}{@{}ll|ccc|ccc|c@{}}
\toprule
&
&
\multicolumn{3}{c|}{\textbf{Chart / Plot VQA}} &
\multicolumn{3}{c|}{\textbf{Medical VQA}} &
\\
\cmidrule(lr){3-5}\cmidrule(lr){6-8}\cmidrule(lr){9-9}
\textbf{Target Model} &
\textbf{BIRD Source} &
\textbf{ChartQA} &
\textbf{ChartBench} &
\textbf{ChartQAPro} &
\textbf{SLAKE} &
\textbf{PathVQA} &
\textbf{MedXpertQA} &
\textbf{Avg.} \\
\midrule

Qwen3.5-9B
& \textbf{Qwen3.5-9B (Own)}
& \textbf{89.84}
& \textbf{71.55}
& \textbf{52.96}
& \textbf{77.92}
& \textbf{51.49}
& \textbf{28.06}
& \textbf{61.97} \\

& InternVL3-8B (Cross)
& 88.05 {\tiny\textcolor{red}{$\downarrow$\,1.79}}
& 69.73 {\tiny\textcolor{red}{$\downarrow$\,1.82}}
& 51.43 {\tiny\textcolor{red}{$\downarrow$\,1.53}}
& 76.42 {\tiny\textcolor{red}{$\downarrow$\,1.50}}
& 49.57 {\tiny\textcolor{red}{$\downarrow$\,1.92}}
& 26.67 {\tiny\textcolor{red}{$\downarrow$\,1.39}}
& 60.31 {\tiny\textcolor{red}{$\downarrow$\,1.66}} \\

\midrule

InternVL3-8B
& \textbf{InternVL3-8B (Own)}
& \textbf{87.06}
& \textbf{69.42}
& \textbf{51.28}
& \textbf{76.31}
& \textbf{51.97}
& \textbf{23.69}
& \textbf{59.96} \\

& Qwen3.5-9B (Cross)
& 85.45 {\tiny\textcolor{red}{$\downarrow$\,1.61}}
& 67.88 {\tiny\textcolor{red}{$\downarrow$\,1.54}}
& 49.14 {\tiny\textcolor{red}{$\downarrow$\,2.14}}
& 74.56 {\tiny\textcolor{red}{$\downarrow$\,1.75}}
& 49.98 {\tiny\textcolor{red}{$\downarrow$\,1.99}}
& 22.34 {\tiny\textcolor{red}{$\downarrow$\,1.35}}
& 58.23 {\tiny\textcolor{red}{$\downarrow$\,1.73}} \\

\bottomrule
\end{tabular}
\vspace{-0.3cm}
\end{table*}

\subsection{Boundary Evidence Analysis}
\label{sec:evidence_analysis}

\begin{figure}
    \centering
    \includegraphics[width=1\linewidth]{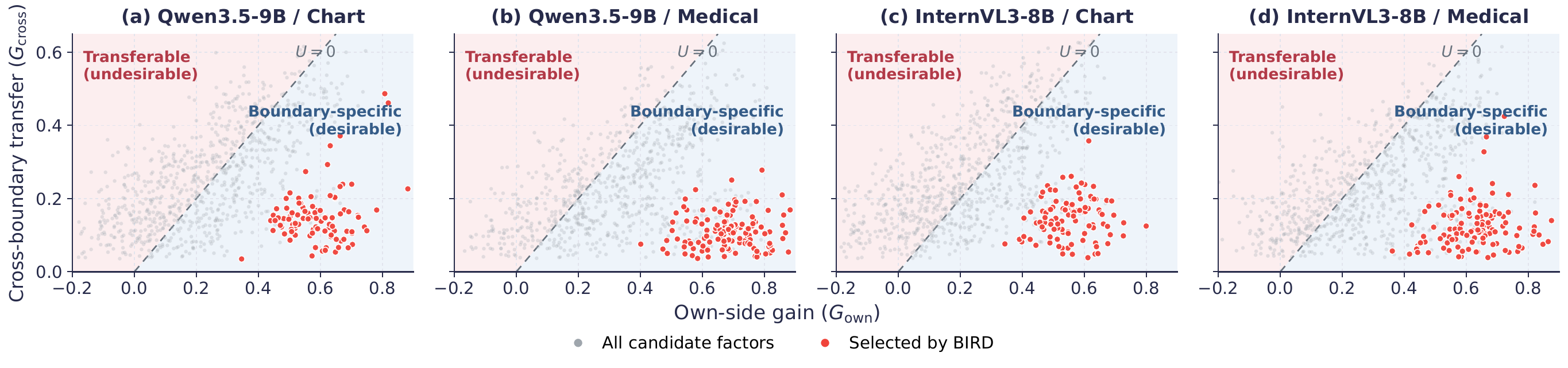}
    \vspace{-0.5cm}
    \caption{Functional distribution of candidate evidence across models and domains. BIRD-selected factors concentrate in the high-own-side-gain, low-cross-boundary-transfer region.}
    \label{fig:fig6}
    \vspace{-0.3cm}
\end{figure}

\textbf{Functional behavior of selected evidence.}
We examine whether BIRD selects factors with the intended boundary
behavior. Figure~\ref{fig:fig6} plots all candidate factors
in the $(G_{\mathrm{own}},G_{\mathrm{cross}})$ space for both target MLLMs and
domains, with BIRD-selected factors highlighted.
Across all four settings, the selected evidence concentrates in the
high-own-side-gain and low-cross-boundary-transfer region, indicating that
boundary utility filters generic pairwise differences into boundary-specific
evidence.

\textbf{Effect of boundary-utility selection.}
We further compare random candidate selection, selecting the factor with the
largest $G_{\mathrm{own}}$, and selecting by the full boundary utility $U$,
while keeping the candidate pool and number of updated rationales fixed.
As shown in Table~\ref{tab:evidence_selection}, selection by $U$ consistently
performs best, showing that maximizing own-side gain alone is insufficient; controlling cross-boundary transfer provides additional benefit.

\begin{table*}[h]
\vspace{-0.1cm}
\centering
\caption{
Effect of different boundary-evidence selection strategies.
Changes are relative to random selection. All variants use the same candidate
pool and update the same number of rationales.
}
\label{tab:evidence_selection}
\scriptsize
\setlength{\tabcolsep}{2.8pt}
\renewcommand{\arraystretch}{1.04}
\begin{tabular}{@{}l|ccc|ccc|c@{}}
\toprule
&
\multicolumn{3}{c|}{\textbf{Chart / Plot VQA}} &
\multicolumn{3}{c|}{\textbf{Medical VQA}} &
\\
\cmidrule(lr){2-4}\cmidrule(lr){5-7}\cmidrule(lr){8-8}
\textbf{Selection}
& \textbf{ChartQA}
& \textbf{ChartBench}
& \textbf{ChartQAPro}
& \textbf{SLAKE}
& \textbf{PathVQA}
& \textbf{MedXpertQA}
& \textbf{Avg.} \\
\midrule

\rowcolor{headergray}\multicolumn{8}{@{}l}{\textit{Qwen3.5-9B}} \\

Random
& 88.71
& 68.82
& 50.21
& 76.12
& 50.04
& 27.74
& 60.27 \\

Max $G_{\mathrm{own}}$
& 89.31 {\tiny\textcolor{blue}{$\uparrow$\,0.60}}
& 69.94 {\tiny\textcolor{blue}{$\uparrow$\,1.12}}
& 51.64 {\tiny\textcolor{blue}{$\uparrow$\,1.43}}
& 76.91 {\tiny\textcolor{blue}{$\uparrow$\,0.79}}
& 50.72 {\tiny\textcolor{blue}{$\uparrow$\,0.68}}
& 27.95 {\tiny\textcolor{blue}{$\uparrow$\,0.21}}
& 61.08 {\tiny\textcolor{blue}{$\uparrow$\,0.81}} \\

\rowcolor{color4}\textbf{Max $U$}
& \textbf{89.84} {\tiny\textcolor{green!60!black}{$\uparrow$\,1.13}}
& \textbf{71.55} {\tiny\textcolor{green!60!black}{$\uparrow$\,2.73}}
& \textbf{52.96} {\tiny\textcolor{green!60!black}{$\uparrow$\,2.75}}
& \textbf{77.92} {\tiny\textcolor{green!60!black}{$\uparrow$\,1.80}}
& \textbf{51.49} {\tiny\textcolor{green!60!black}{$\uparrow$\,1.45}}
& \textbf{28.06} {\tiny\textcolor{green!60!black}{$\uparrow$\,0.32}}
& \textbf{61.97} {\tiny\textcolor{green!60!black}{$\uparrow$\,1.70}} \\

\midrule

\rowcolor{headergray}\multicolumn{8}{@{}l}{\textit{InternVL3-8B}} \\

Random
& 85.73
& 66.92
& 48.52
& 74.62
& 50.41
& 23.24
& 58.24 \\

Max $G_{\mathrm{own}}$
& 86.34 {\tiny\textcolor{blue}{$\uparrow$\,0.61}}
& 68.06 {\tiny\textcolor{blue}{$\uparrow$\,1.14}}
& 49.97 {\tiny\textcolor{blue}{$\uparrow$\,1.45}}
& 75.38 {\tiny\textcolor{blue}{$\uparrow$\,0.76}}
& 51.05 {\tiny\textcolor{blue}{$\uparrow$\,0.64}}
& 23.47 {\tiny\textcolor{blue}{$\uparrow$\,0.23}}
& 59.04 {\tiny\textcolor{blue}{$\uparrow$\,0.80}} \\

\rowcolor{color4}\textbf{Max $U$}
& \textbf{87.06} {\tiny\textcolor{green!60!black}{$\uparrow$\,1.33}}
& \textbf{69.42} {\tiny\textcolor{green!60!black}{$\uparrow$\,2.50}}
& \textbf{51.28} {\tiny\textcolor{green!60!black}{$\uparrow$\,2.76}}
& \textbf{76.31} {\tiny\textcolor{green!60!black}{$\uparrow$\,1.69}}
& \textbf{51.97} {\tiny\textcolor{green!60!black}{$\uparrow$\,1.56}}
& \textbf{23.69} {\tiny\textcolor{green!60!black}{$\uparrow$\,0.45}}
& \textbf{59.96} {\tiny\textcolor{green!60!black}{$\uparrow$\,1.72}} \\

\bottomrule
\end{tabular}
\vspace{-0.3cm}
\end{table*}

\subsection{Ablation Study}
\label{sec:ablation}


We ablate two components as shown in Table~\ref{tab:ablation}.

\textbf{Effect of Confuser Selection.}
We remove explicit confuser selection by directly using the alternative associated with the nearest valid neighbor, rather than selecting the lowest-energy alternative of the target MLLM. As shown in Table~\ref{tab:ablation}, this variant consistently degrades performance, indicating that confuser selection better targets the model's unresolved ambiguity.

\textbf{Effect of Rationale Distillation.}
We remove rationale distillation by appending the verified target evidence to the original rationale without target-conditioned refinement. The resulting performance drop in Table~\ref{tab:ablation} demonstrates that rationale distillation more effectively converts pairwise boundary evidence into sample-specific supervision.

\begin{table*}[h]
\centering
\vspace{-0.3cm}
\caption{
Ablation of confuser selection and rationale distillation.
Arrows indicate absolute performance drops relative to full BIRD.
}
\label{tab:ablation}
\scriptsize
\setlength{\tabcolsep}{2.8pt}
\renewcommand{\arraystretch}{1.04}
\begin{tabular}{@{}l|ccc|ccc|c@{}}
\toprule
&
\multicolumn{3}{c|}{\textbf{Chart / Plot VQA}} &
\multicolumn{3}{c|}{\textbf{Medical VQA}} \\
\cmidrule(lr){2-4}\cmidrule(lr){5-7}\cmidrule(lr){8-8}
\textbf{Variant}
& \textbf{ChartQA}
& \textbf{ChartBench}
& \textbf{ChartQAPro}
& \textbf{SLAKE}
& \textbf{PathVQA}
& \textbf{MedXpertQA}
& \textbf{Avg.} \\
\midrule

\rowcolor{headergray}\multicolumn{8}{@{}l}{\textit{Qwen3.5-9B}} \\

w/o Confuser Selection
& 89.50 {\tiny\textcolor{red}{$\downarrow$\,0.34}}
& 69.45 {\tiny\textcolor{red}{$\downarrow$\,2.10}}
& 51.55 {\tiny\textcolor{red}{$\downarrow$\,1.41}}
& 76.18 {\tiny\textcolor{red}{$\downarrow$\,1.74}}
& 50.50 {\tiny\textcolor{red}{$\downarrow$\,0.99}}
& 27.92 {\tiny\textcolor{red}{$\downarrow$\,0.14}}
& 60.88 {\tiny\textcolor{red}{$\downarrow$\,1.09}} \\

w/o Rationale Distillation
& 89.66 {\tiny\textcolor{red}{$\downarrow$\,0.18}}
& 70.88 {\tiny\textcolor{red}{$\downarrow$\,0.67}}
& 52.43 {\tiny\textcolor{red}{$\downarrow$\,0.53}}
& 77.31 {\tiny\textcolor{red}{$\downarrow$\,0.61}}
& 51.12 {\tiny\textcolor{red}{$\downarrow$\,0.37}}
& 27.97 {\tiny\textcolor{red}{$\downarrow$\,0.09}}
& 61.59 {\tiny\textcolor{red}{$\downarrow$\,0.38}} \\

\rowcolor{color4}\textbf{BIRD}
& \textbf{89.84}
& \textbf{71.55}
& \textbf{52.96}
& \textbf{77.92}
& \textbf{51.49}
& \textbf{28.06}
& \textbf{61.97} \\

\midrule

\rowcolor{headergray}\multicolumn{8}{@{}l}{\textit{InternVL3-8B}} \\

w/o Confuser Selection
& 86.45 {\tiny\textcolor{red}{$\downarrow$\,0.61}}
& 67.80 {\tiny\textcolor{red}{$\downarrow$\,1.62}}
& 49.88 {\tiny\textcolor{red}{$\downarrow$\,1.40}}
& 75.02 {\tiny\textcolor{red}{$\downarrow$\,1.29}}
& 51.02 {\tiny\textcolor{red}{$\downarrow$\,0.95}}
& 23.05 {\tiny\textcolor{red}{$\downarrow$\,0.64}}
& 58.99 {\tiny\textcolor{red}{$\downarrow$\,0.97}} \\

w/o Rationale Distillation
& 86.78 {\tiny\textcolor{red}{$\downarrow$\,0.28}}
& 68.85 {\tiny\textcolor{red}{$\downarrow$\,0.57}}
& 50.82 {\tiny\textcolor{red}{$\downarrow$\,0.46}}
& 75.82 {\tiny\textcolor{red}{$\downarrow$\,0.49}}
& 51.55 {\tiny\textcolor{red}{$\downarrow$\,0.42}}
& 23.60 {\tiny\textcolor{red}{$\downarrow$\,0.09}}
& 59.60 {\tiny\textcolor{red}{$\downarrow$\,0.36}} \\

\rowcolor{color4}\textbf{BIRD}
& \textbf{87.06}
& \textbf{69.42}
& \textbf{51.28}
& \textbf{76.31}
& \textbf{51.97}
& \textbf{23.69}
& \textbf{59.96} \\

\bottomrule
\end{tabular}
\vspace{-0.3cm}
\end{table*}

\section{Conclusion}


We present BIRD, a self-improving framework that turns the target MLLM’s own confusions into rationale supervision. BIRD retrieves model-specific candidate neighbors, selects the alternative the model finds most confusable, and generates answer-blind candidate factors from the resulting pair. It retains factors that improve answer separation on their corresponding samples without transferring across the pair, and distills the verified target-side evidence into a single-sample rationale. Experiments on medical and chart VQA show that BIRD achieves the strongest average performance across both target MLLMs. Further analyses demonstrate greater separation on held-out hard pairs and stronger gains from model-matched supervision. Overall, our findings establish decision boundaries as a useful source of self-discovered supervision and offer a promising direction for adapting MLLMs to specialized domains.

\section*{AI Use Statement}

AI-assisted tools were used to improve the language and clarity of portions of this manuscript, generate synthetic datasets, and implement parts of the code under the direct guidance and supervision of the authors. All AI-assisted code was subsequently reviewed and validated by two authors, who examined its implementation logic and verified its correctness and consistency with the intended methodology.

\section*{Ethics Statement}

This work uses publicly released research datasets and introduces no new human-subject data collection. Because our experiments include medical VQA, the resulting models and rationales should not be interpreted as clinically reliable or used for diagnosis or treatment. Model-generated rationales may inherit biases or contain unsupported statements, and therefore require appropriate human oversight in high-stakes applications.

\section*{Reproducibility Statement}

We provide dataset and training details, hyperparameters, the complete BIRD algorithm, and all prompt templates in the Appendix. Evaluation protocols and construction statistics are also reported.

\bibliography{iclr2027_conference}
\bibliographystyle{iclr2027_conference}

\newpage
\appendix
\section*{Appendix}
\section{Experimental Details}
\label{app:experimental_details}

\subsection{Datasets}
\label{app:datasets}

We consider two domain-adaptation settings, chart/plot VQA and medical VQA.
ChartQA-X and OpenMedReason serve as the source datasets for domain adaptation,
while the remaining datasets are used as held-out benchmarks.

\paragraph{Chart / Plot Domain.}
\textbf{ChartQA-X} extends ChartQA with natural-language explanations for chart question–answer pairs. Its training split contains 28,299 examples with model-generated explanations. We use this split as the source data for chart-domain adaptation.

\textbf{ChartQA} contains real-world charts paired with human-authored questions and questions generated from chart summaries. We evaluate on its official test split of 2,500 examples.

\textbf{ChartBench} substantially broadens the
visual diversity of chart reasoning, covering nine major chart types and
42 fine-grained categories. Its training corpus contains approximately
66.6K charts and 599.6K QA pairs, while the held-out benchmark contains
2,100 charts and 18,900 QA pairs. Many charts omit explicit data-point
annotations, requiring reasoning directly from visual elements such as
axes, legends, colors, and graphical marks.

\textbf{ChartQAPro} targets more diverse and
challenging real-world chart understanding. It contains 1,341 charts
collected from 157 sources and 1,948 questions, spanning conventional
charts as well as infographics and dashboards. The questions cover
multiple formats, including factoid, multiple-choice, conversational,
hypothetical, and unanswerable cases.

\paragraph{Medical Domain.}
\textbf{OpenMedReason} is a medical VQA dataset constructed from figures in open-access biomedical literature. Each example pairs an image with a multiple-choice question, an answer, and an image-grounded reasoning trace. The released training split contains 150,246 examples, which we use as the source data for medical-domain adaptation.

\textbf{SLAKE} is a physician-annotated bilingual
medical VQA dataset built from 642 radiology images, including CT, MRI,
and X-ray scans. The complete bilingual dataset contains 14,028 QA pairs.
Following common English-language evaluation settings, we use its English
subset, consisting of 4,919 training, 1,053 validation, and 1,061 test
questions.

\textbf{PathVQA} focuses on pathology images and
contains approximately 32.8K QA pairs over 4,998 images. It includes both
open-ended questions and closed-ended yes/no questions covering visual
properties such as location, appearance, shape, and color. The standard
split used in our experiments contains 19,755 training, 6,279 validation,
and 6,761 test QA pairs.

\textbf{MedXpertQA} is designed to evaluate expert-level medical understanding
and reasoning, and we use its multimodal subset. It spans 17 medical specialties and 11 body systems and
combines clinical context with one or more medical images. The released
multimodal subset contains 5 development examples and 2,000 test
questions, which constitute the benchmark used in our experiments.

\subsection{Fine-tuning Details}
\label{app:finetuning}

We perform full-parameter supervised fine-tuning. Unless otherwise specified, the vision encoder, multimodal alignment modules, and language model are all trainable. All methods use the same source training data and target MLLM within each model--domain setting, and the same
SFT configuration is used whenever applicable.

Table~\ref{tab:training_hyperparameters} reports the details. Our training and inference are both built on the ms-swift framework~\citep{zhao2024swiftascalablelightweightinfrastructure}. We train each model for 10 epochs on 8 NVIDIA A800 GPUs using AdamW with a learning rate of $2\times10^{-5}$, weight decay of $0.1$, and $(\beta_1,\beta_2)=(0.9,0.95)$. We use a cosine learning-rate schedule with a $3\%$ warmup ratio and BF16 precision. The per-device batch size is 4 with 4 gradient-accumulation steps, resulting in a global batch size of 128.
The maximum sequence length is 1024. Training uses DeepSpeed ZeRO-3 and FlashAttention.

\begin{table}[h]
\centering
\caption{Training hyperparameters used for all model--domain settings.}
\label{tab:training_hyperparameters}
\small
\setlength{\tabcolsep}{6pt}
\begin{tabular}{@{}ll@{}}
\toprule
\textbf{Hyperparameter} & \textbf{Value} \\
\midrule
Fine-tuning type & Full-parameter SFT \\
Training epochs & 10 \\
Optimizer & AdamW \\
Learning rate & $2\times10^{-5}$ \\
Weight decay & $0.1$ \\
Adam $(\beta_1,\beta_2)$ & $(0.9,0.95)$ \\
LR schedule & Cosine \\
Warmup ratio & $3\%$ \\
Precision & BF16 \\
Per-device batch size & 4 \\
Gradient accumulation & 4 \\
Global batch size & 128 \\
Maximum sequence length & 1024 \\
Gradient checkpointing & Yes \\
Distributed training & DeepSpeed ZeRO-3 \\
Attention implementation & FlashAttention \\
Training hardware & 8 NVIDIA A800 GPUs \\
\bottomrule
\end{tabular}
\end{table}

\section{BIRD Construction Details}
\label{app:bird_details}

This section provides implementation details omitted from the main text.
Algorithm~\ref{alg:bird} summarizes the complete offline construction
procedure, followed by the prompt templates used in BIRD.

\subsection{End-to-End Algorithm}
\label{app:bird_algorithm}
We clearly show the pseudo code of BIRD, as shown in Algorithm~\ref{alg:bird}.
\begin{algorithm}[h]
\caption{Boundary-Informed Rationale Distillation (BIRD)}
\label{alg:bird}
\small
\begin{algorithmic}[1]
\Require Training set $\mathcal{D}_0$; target MLLM $M_\theta$;
retrieval size $P$; number of candidate factors $K$;
cross-boundary penalty $\lambda$; utility threshold $\tau_U$
\Ensure Boundary-informed training set $\mathcal{D}^{*}$

\State Extract and cache visual representations
$\{\mathbf{v}_i\}_{i=1}^{N}$
\State $\mathcal{D}^{*} \gets \emptyset$

\For{each $(I_i,Q_i,R_i^0,A_i)\in\mathcal{D}_0$}
    \State $R_i^{*}\gets R_i^0$
    \State Construct a question-compatible pool $\mathcal{P}_i$
    \State Map each valid candidate answer into the answer space of $Q_i$
    \State Retrieve the top-$P$ visual neighbors
    $\mathcal{N}_i\subseteq\mathcal{P}_i$

    \If{$\mathcal{N}_i\neq\emptyset$}
        \State Form the alternative-answer set
        $\mathcal{A}_i^{-}$ from $\mathcal{N}_i$
        \State $C_i\gets
        \arg\min_{c\in\mathcal{A}_i^{-}}
        E_\theta(c\mid I_i,Q_i)$
        \State Select the closest neighbor supporting $C_i$
        as the witness image $I_i^{-}$

        \State Use $M_\theta$ to generate $K$ answer-blind factor pairs
        \[
        \mathcal{B}_i
        =
        \{b_i^k=(e_i^k,e_{i^-}^k)\}_{k=1}^{K}
        \]

        \If{$\mathcal{B}_i\neq\emptyset$}
            \For{each $b_i^k\in\mathcal{B}_i$}
                \State Compute $G_{i,\mathrm{own}}^k$
                and $G_{i,\mathrm{cross}}^k$
                \State $U_i^k\gets
                G_{i,\mathrm{own}}^k
                -\lambda G_{i,\mathrm{cross}}^k$
            \EndFor

            \State $k^{*}\gets
            \arg\max_{k:\,b_i^k\in\mathcal{B}_i}U_i^k$

            \If{$U_i^{k^{*}}\geq\tau_U$}
                \State $R_i^{*}\gets
                \mathrm{Refine}\!\left(
                R_i^0;I_i,Q_i,A_i,C_i,
                e_i^{k^{*}},e_{i^-}^{k^{*}}
                \right)$
            \EndIf
        \EndIf
    \EndIf

    \State $\mathcal{D}^{*}\gets
    \mathcal{D}^{*}\cup\{(I_i,Q_i,R_i^{*},A_i)\}$
\EndFor

\State \Return $\mathcal{D}^{*}$
\end{algorithmic}
\end{algorithm}

\subsection{Question-Compatible Neighbor Pool Construction}
\label{app:compatible_pool}

For datasets with recurring question templates, we group samples by their
normalized template before visual retrieval. Otherwise, we use Prompt~P1 to
verify that two questions concern the same underlying visual property and to
map the candidate QA state into the answer space of the target question.
Candidates with ambiguous mappings or with the same mapped answer as the
target are discarded. Top-$P$ visual retrieval is then performed within this
compatible pool using the target MLLM representation defined in the main text.

\subsection{Answer-Blind Candidate Evidence Generation}
\label{app:evidence_generation}

Given the selected target--witness pair, Prompt~P2 receives only the two
images and the target question; the gold and confuser answers are withheld.
The target MLLM produces $K$ paired observations
$(e_i^k,e_{i^-}^k)$, each describing the same question-relevant visual factor
on the two sides. Candidates that explicitly reveal an answer are removed
before scoring. No candidate is treated as valid boundary evidence until it
passes the functional verification step.

\subsection{Boundary Utility Computation}
\label{app:utility_computation}

We compute $G_{\mathrm{own}}$, $G_{\mathrm{cross}}$, and $U$ exactly as
defined in Eqs.~(11)--(17) of the main text. Answer energies are obtained by
teacher-forcing each candidate answer and averaging its token-level negative
log-likelihood. For a given pair, the evidence-free energy gaps are cached
once and reused across candidate factors. Matched and swapped evaluations use
the same input format; only the inserted evidence string is changed.
The factor with the largest $U$ is retained, and rationale refinement is
performed only if its utility exceeds $\tau_U$.

\subsection{Rationale Distillation}
\label{app:rationale_distillation}

For a verified factor, Prompt~P3 refines the original rationale using the
target-side observation, while the witness-side observation is provided only
as contrastive context. The output must remain a rationale for the target
image alone: valid content from the original rationale is preserved, whereas
unsupported details and explicit references to the neighboring sample or the
BIRD construction process are excluded. Samples below the utility threshold
retain their original rationales.

\subsection{Full Prompt Templates}
\label{app:bird_prompts}

Text enclosed in braces denotes an instance-specific field.

\begin{promptbox}{Prompt P1: Question-Compatible Neighbor Validation}[t]

\prompttask{
Determine whether the candidate QA sample can serve as a valid contrastive
neighbor for the target question.
}

\promptinput{Target question:}\\
\slot{target\_question}

\vspace{1mm}
\promptinput{Target answer:}\\
\slot{target\_answer}

\vspace{1mm}
\promptinput{Candidate question:}\\
\slot{candidate\_question}

\vspace{1mm}
\promptinput{Candidate answer:}\\
\slot{candidate\_answer}

\vspace{2mm}
\promptconstraint{Criteria.}

A candidate is valid only if:
\begin{enumerate}
    \item both questions concern the same underlying visual property;
    \item the candidate QA state can be unambiguously expressed as an
    answer to the target question; and
    \item the mapped answer differs from the target answer.
\end{enumerate}

Do not judge visual similarity.

\promptoutput{
Return \texttt{Decision: VALID} or \texttt{Decision: INVALID}.
If VALID, additionally return
\texttt{Mapped answer: <answer>}.
}

\end{promptbox}

\begin{promptbox}{Prompt P2: Answer-Blind Candidate Evidence Generation}

\prompttask{
Compare the two images under the given question and identify visual
differences that could distinguish them.
}

\promptinput{Question:}\\
\slot{question}

\vspace{2mm}
\promptconstraint{Constraints.}

\begin{enumerate}
    \item Each factor must contain corresponding observations for both images.
    \item Include only differences relevant to the given question.
    \item Do not use wording that explicitly reveals an answer.
\end{enumerate}

Generate up to \slot{num\_factors} distinct factors.

\promptoutput{
Use the following format:
}

\begin{quote}
\ttfamily\small
Factor 1\\
Image A: <observation in the target image>\\
Image B: <corresponding observation in the witness image>

\vspace{1mm}
Factor 2\\
Image A: ...\\
Image B: ...
\end{quote}

\end{promptbox}

\begin{promptbox}{Prompt P3: Boundary-Informed Rationale Distillation}

\prompttask{
Refine the original rationale using the verified discriminative factor.
}

\promptinput{Question:}\\
\slot{question}

\vspace{1mm}
\promptinput{Correct answer:}\\
\slot{target\_answer}

\vspace{1mm}
\promptinput{Confusable alternative:}\\
\slot{confuser}

\vspace{1mm}
\promptinput{Original rationale:}\\
\slot{original\_rationale}

\vspace{1mm}
\promptinput{Verified target-side observation:}\\
\slot{target\_evidence}

\vspace{1mm}
\promptinput{Corresponding neighbor-side observation:}\\
\slot{neighbor\_evidence}

\vspace{2mm}
\promptconstraint{Requirements.}

\begin{enumerate}
    \item Preserve correct and useful content from the original rationale.
    \item Incorporate the verified target-side observation.
    \item Do not mention the neighboring sample.
    \item Remove unsupported details from the original rationale and do not introduce new visual observations that are not supported by the target image.
\end{enumerate}

\promptoutput{
Return only the refined rationale for the target image.
}

\end{promptbox}

\section{Additional Analysis}
\label{app:additional_analysis}

\subsection{Per-Benchmark Boundary Sharpening}
\label{app:per_benchmark_boundary}

The main text reports boundary separation aggregated within each
model--domain setting. Here, we further break down the analysis by benchmark
to examine whether the observed boundary sharpening is consistent across
individual evaluation sets.

\begin{figure}[h]
    \centering
    \includegraphics[width=1\linewidth]{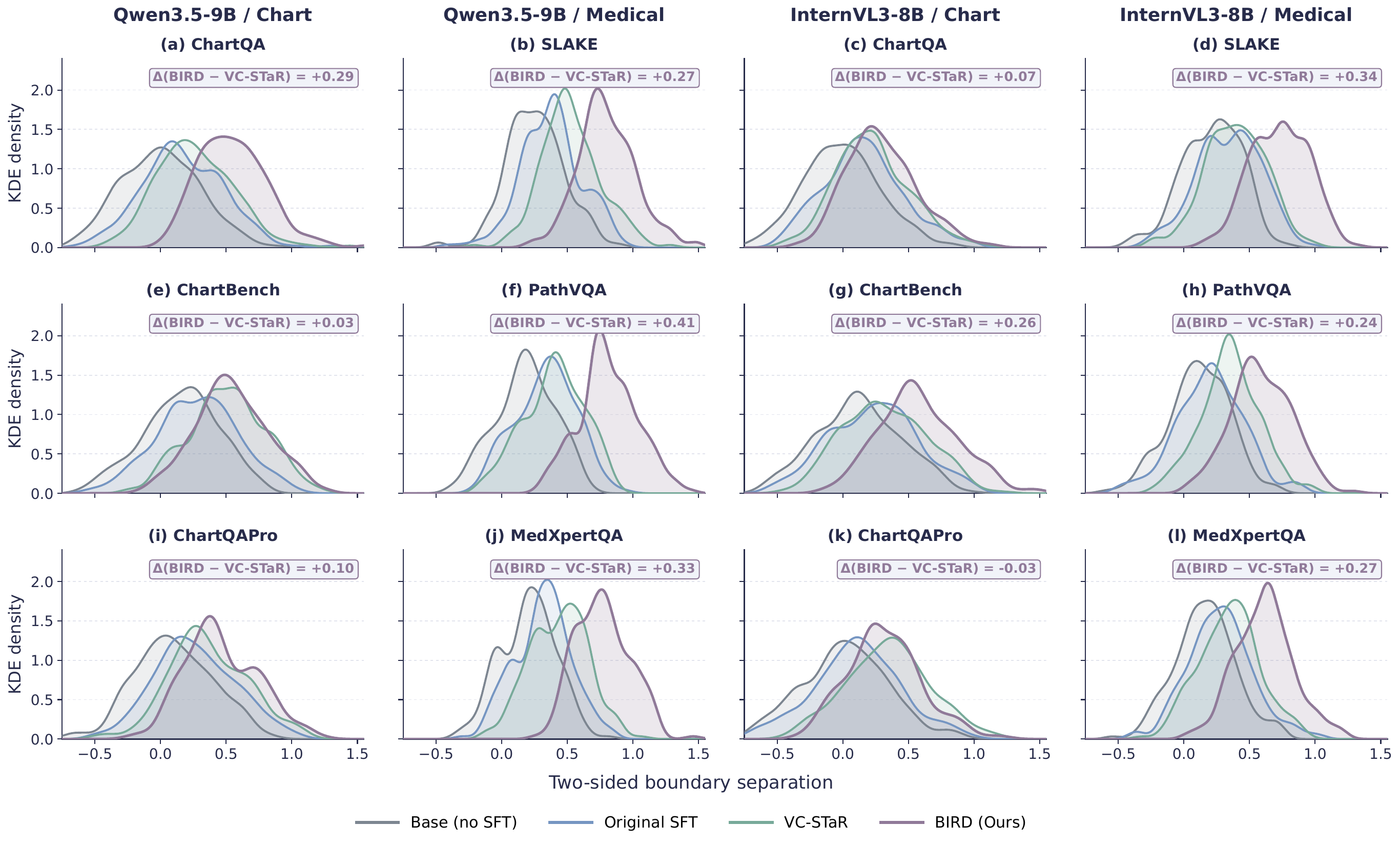}
    \vspace{-0.2cm}
    \caption{Per-benchmark analysis of decision-boundary sharpening.
    Each panel shows the distribution of two-sided boundary separation
    $S_i$ on 50 fixed held-out hard pairs. Larger values indicate clearer
    separation between the gold answer and the pre-adaptation confuser.}
    \label{fig:fig7}
\end{figure}

We use the same fixed hard pairs constructed using Kimi K3 as in Sec.~\ref{sec:boundary_sharpening}, with 50 pairs sampled from each benchmark. All methods are evaluated on exactly the same pairs. Figure~\ref{fig:fig7}
reports the distribution of the two-sided boundary separation $S_i$ for
Base, Original SFT, VC-STaR, and BIRD on each benchmark separately.

Across both target MLLMs, BIRD improves the mean separation in 11 of the 12 model–benchmark combinations and is nearly tied with VC-STaR on InternVL3-8B/ChartQAPro. The improvements are generally larger on the medical benchmarks. These results show that the domain-level boundary sharpening reported in the main text is not driven by a particular benchmark.

\subsection{Rationale Update Statistics}
\label{app:rationale_statistics}

BIRD updates a training rationale only when a valid confusable pair can be
formed, candidate evidence survives answer-leakage filtering, and the best
factor satisfies $U^{k^*}\geq\tau$. Table~\ref{tab:rationale_statistics}
summarizes the resulting construction statistics for each target
model--domain setting.

We report three coverage statistics: \emph{pair coverage}, the fraction of
training samples for which a valid question-compatible different-answer
witness is obtained; \emph{candidate coverage}, the fraction for which at
least one answer-blind candidate remains after filtering; and
\emph{update rate}, the fraction whose rationale is ultimately refined.
For updated samples, we additionally report the mean own-side gain,
cross-boundary transfer, and boundary utility of the selected factor.

\begin{table}[h]
\centering
\caption{
Statistics of BIRD rationale construction.
Pair and candidate coverage, and update rate are percentages over the
source training set. $G_{\mathrm{own}}$, $G_{\mathrm{cross}}$, and $U$
are averaged over samples whose rationales are updated.
}
\label{tab:rationale_statistics}
\scriptsize
\setlength{\tabcolsep}{4.0pt}
\renewcommand{\arraystretch}{1.08}
\begin{tabular}{@{}llccc|ccc@{}}
\toprule
\textbf{Model} &
\textbf{Domain} &
\textbf{Pair} &
\textbf{Candidate} &
\textbf{Update} &
$\mathbf{G_{\mathrm{own}}}$ &
$\mathbf{G_{\mathrm{cross}}}$ &
$\mathbf{U}$ \\
&
&
\textbf{Cov. (\%)} &
\textbf{Cov. (\%)} &
\textbf{Rate (\%)} &
&
&
\\
\midrule
Qwen3.5-9B
& Chart / Plot & 93.8 & 89.7 & 70.9 & 1.25 & 0.17 & 0.91 \\
& Medical      & 88.4 & 84.2 & 67.8 & 1.61 & 0.15 & 1.31 \\
\midrule
InternVL3-8B
& Chart / Plot & 92.9 & 88.6 & 68.6 & 1.14 & 0.15 & 0.84 \\
& Medical      & 87.6 & 82.9 & 65.7 & 1.46 & 0.15 & 1.16 \\
\bottomrule
\end{tabular}
\vspace{-0.2cm}
\end{table}

BIRD forms valid comparison pairs for most training samples, while
candidate filtering and boundary-utility thresholding make rationale
updates more selective. With the default threshold $\tau=0.2$,
$65.7\%$--$70.9\%$ of the training rationales are refined. The selected
factors consistently exhibit substantially larger own-side gains than
cross-boundary transfer under the default penalty $\lambda=2.0$.

\subsection{Rationale Length Analysis}
\label{sec:rationale_length}

A possible concern is that BIRD may improve adaptation simply by producing longer rationales and thus providing more supervision tokens during fine-tuning.
\begin{wrapfigure}[15]{r}{0.5\columnwidth}
    \centering
    \includegraphics[width=\linewidth]{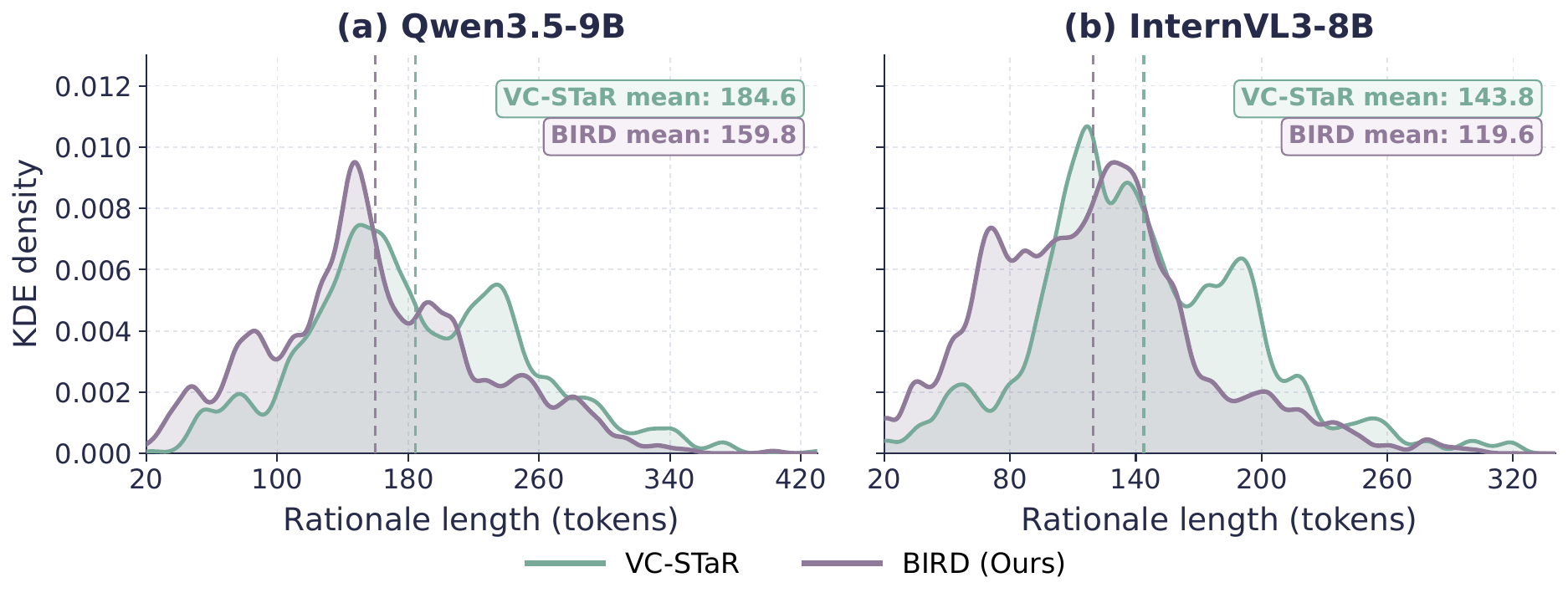}
    \vspace{-0.5cm}
    \caption{Rationale-length distributions of VC-STaR and BIRD for (a) Qwen3.5-9B and (b) InternVL3-8B. This indicates that BIRD does not obtain richer supervision simply by increasing rationale verbosity.}
    \label{fig:fig8}
\end{wrapfigure}
To test this, we compare the rationale-length distributions of VC-STaR and BIRD for both target MLLMs in Figure~\ref{fig:fig8}. The distributions substantially overlap and fall within similar overall length ranges, indicating comparable rationale budgets. Notably, BIRD is not shifted toward longer rationales: its average rationale length is 159.8 tokens versus 184.6 for VC-STaR on Qwen3.5-9B, and 119.6 versus 143.8 on InternVL3-8B. The density profiles show a consistent pattern: BIRD retains a broad distribution comparable to VC-STaR but places slightly more mass in the shorter-length region, despite differences in the models’ absolute rationale lengths. Thus, BIRD’s additional supervision does not arise from increased verbosity or more rationale tokens. Instead, it changes the content of supervision by selectively incorporating verified boundary-relevant evidence within a comparable, and on average smaller, rationale budget. Therefore, BIRD’s advantage cannot be readily attributed to a larger amount of textual supervision.

\section{Sensitivity Analysis}
\label{app:sensitivity}

\begin{figure}[h]
\centering
\includegraphics[width=1\linewidth]{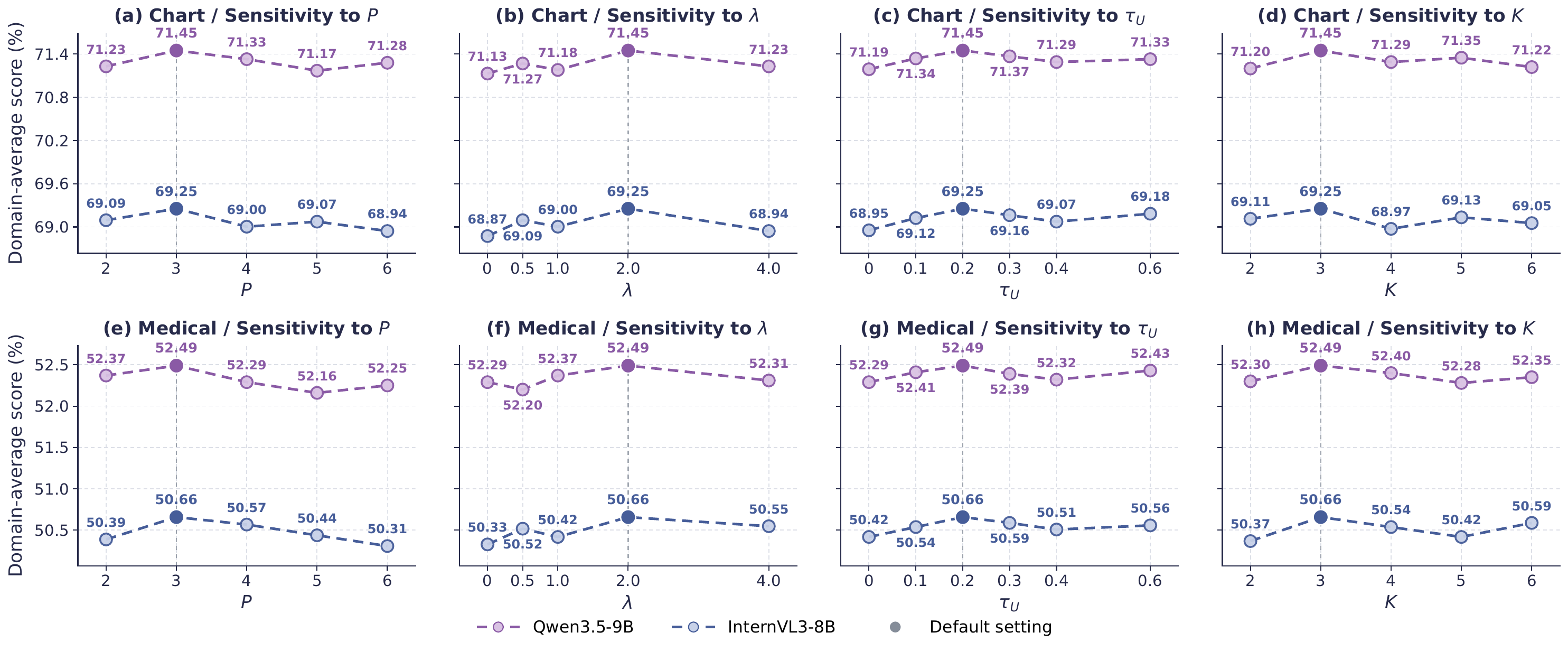}
\vspace{-0.3cm}
\caption{Sensitivity of BIRD to the retrieval size $P$, cross-boundary
penalty $\lambda$, utility threshold $\tau_U$, and number of candidate
factors $K$. The top and bottom rows report results on the chart and
medical domains, respectively. Scores are averaged over the three
benchmarks in each domain. Filled markers indicate the default settings
($P=3$, $\lambda=2.0$, $\tau_U=0.2$, and $K=3$), which consistently
achieve the strongest performance across both target MLLMs and domains.}
\vspace{-0.3cm}
\label{fig:fig9}
\end{figure}

We examine the sensitivity of BIRD to four construction hyperparameters in
Figure~\ref{fig:fig9}: the retrieval size $P$, cross-boundary penalty
$\lambda$, utility threshold $\tau_U$, and number of candidate factors $K$.
We vary one hyperparameter at a time while keeping the others fixed at their
default values. For each model--domain setting, performance is averaged over
the three corresponding benchmarks. Performance varies by less than $0.4$
percentage points within every sweep, demonstrating that BIRD is generally
robust to these hyperparameters. Moreover, the selected default consistently
achieves the highest score across both target MLLMs and domains.

\subsection{Sensitivity to the Retrieval Size $P$}
\label{app:sensitivity_p}

The retrieval size $P$ controls how many visually similar,
question-compatible neighbors are retained before BIRD identifies the
alternative most confusable to the target MLLM. We vary
$P\in{2,3,4,5,6}$ while keeping the other settings fixed at their default
values.

Increasing $P$ from $2$ to $3$ improves performance in every model--domain
setting. Further enlarging the retrieval pool provides no consistent
benefit and generally leads to small declines. This suggests that a compact
neighborhood already captures the most informative unresolved confusions,
whereas a larger pool may introduce more distant or less relevant
alternatives. We therefore set $P=3$.

\subsection{Sensitivity to the Cross-Boundary Penalty $\lambda$}
\label{app:sensitivity_lambda}

The coefficient $\lambda$ controls how strongly BIRD penalizes
cross-boundary transfer when evaluating a candidate factor. We vary
$\lambda\in{0,0.5,1.0,2.0,4.0}$ while keeping the other settings fixed at
their default values.

Compared with $\lambda=0$, introducing a moderate cross-boundary penalty
generally improves performance, with $\lambda=2.0$ achieving the highest
score in every model--domain setting. Increasing the penalty further to
$\lambda=4.0$ consistently reduces performance. These results confirm the
importance of controlling cross-boundary transfer, while suggesting that an
overly strong penalty may suppress factors that provide useful own-side
gains despite limited transfer across the boundary.

\subsection{Sensitivity to the Utility Threshold $\tau_U$}
\label{app:sensitivity_tau}

The threshold $\tau_U$ determines whether the highest-utility factor is
sufficiently boundary-relevant to trigger rationale refinement. We vary
$\tau_U\in{0,0.1,0.2,0.3,0.4,0.6}$ while keeping the other settings fixed
at their default values.

Performance improves as $\tau_U$ increases from $0$ to $0.2$, indicating
that filtering weak candidate factors benefits rationale refinement. The
default $\tau_U=0.2$ achieves the highest score across all model--domain
settings, while both lower and higher thresholds produce only modest
decreases. This pattern reflects a balance between admitting weak evidence
and rejecting potentially useful rationale updates.

\subsection{Sensitivity to the Number of Candidate Factors $K$}
\label{app:sensitivity_k}

The number $K$ controls how many answer-blind factor pairs are generated
from each target--witness pair before answer-leakage filtering and
boundary-utility evaluation. We vary $K\in{2,3,4,5,6}$ while keeping the
other settings fixed at their default values.

Increasing $K$ from $2$ to $3$ improves performance across both target
MLLMs and domains. Beyond $K=3$, the scores fluctuate slightly but remain
consistently below the default. This suggests that a small candidate set is
sufficient to cover the principal distinctions between paired samples,
whereas additional factors are more likely to be redundant or less
informative.

\section{Case Studies}
\label{app:case_studies}

We provide four case studies to illustrate how BIRD converts model-specific
confusions into boundary-informed rationale supervision. Figures~\ref{fig:fig10}
and~\ref{fig:fig11} present complete examples from chart and medical VQA,
respectively. Figure~\ref{fig:fig12} explains why a factor with a large
target-side gain may still fail to capture the relevant decision boundary.
Figure~\ref{fig:fig13} further shows that both the confusable
alternative and the evidence needed to resolve it depend on the target MLLM.

\subsection{Chart/Plot VQA Examples}
\label{app:case_chart}

\begin{figure}[t]
\centering
\includegraphics[width=\linewidth]{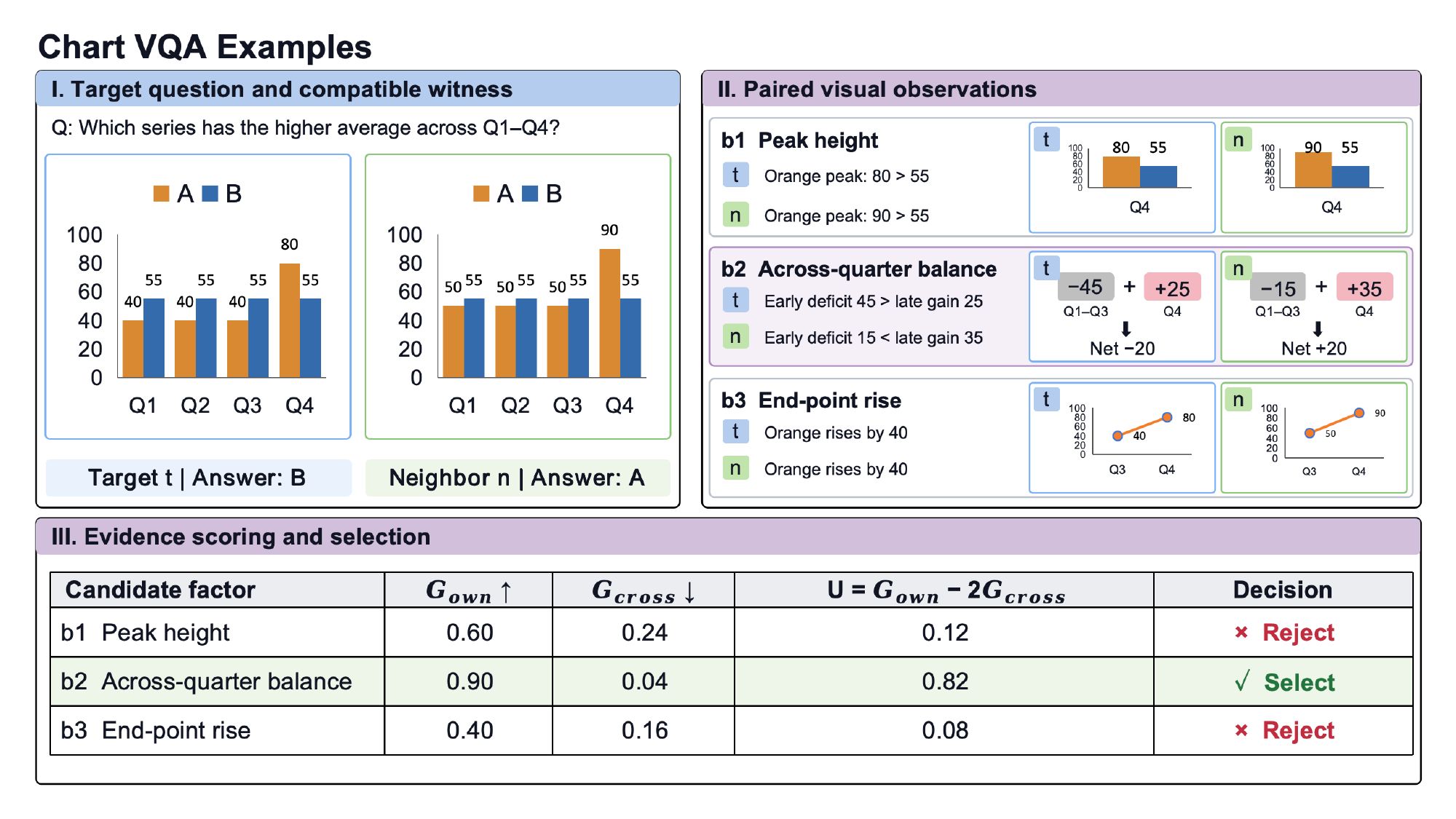}
\caption{Chart VQA example. BIRD compares a target answered B with a
compatible witness answered A and evaluates three answer-blind paired
factors. Across-quarter balance (b2) is selected because it provides high
own-side gain and little cross-boundary transfer, whereas peak height (b1)
and end-point rise (b3) describe patterns shared by both charts.}
\label{fig:fig10}
\end{figure}

Figure~\ref{fig:fig10} shows a chart question asking which series has the
higher average across four quarters. The target sample has answer B, whereas
the retrieved witness has answer A. Although the two charts share similar
colors, scales, and overall structures, the balance between the early quarters
and Q4 reverses the correct answer.

BIRD generates three answer-blind paired factors. Peak height (b1) and
end-point rise (b3) both describe visible patterns, but these patterns occur
on both sides of the pair and therefore do not explain the answer reversal.
Their cross-boundary transfer reduces their utilities to $0.12$ and $0.08$,
respectively. In contrast, the across-quarter balance (b2) captures the
decisive difference. In the target, A's cumulative deficit of $45$ over
Q1--Q3 exceeds its gain of $25$ in Q4, leaving B ahead overall. In the
witness, the early deficit is only $15$, while the Q4 gain is $35$, reversing
the result. This factor achieves $G_{\mathrm{own}}=0.90$ and
$G_{\mathrm{cross}}=0.04$, giving the highest utility of $0.82$. BIRD
therefore distills this cumulative comparison, rather than the visually
salient but non-decisive Q4 peak, into the target rationale.

\subsection{Medical VQA Examples}
\label{app:case_medical}

\begin{figure}[t]
\centering
\includegraphics[width=\linewidth]{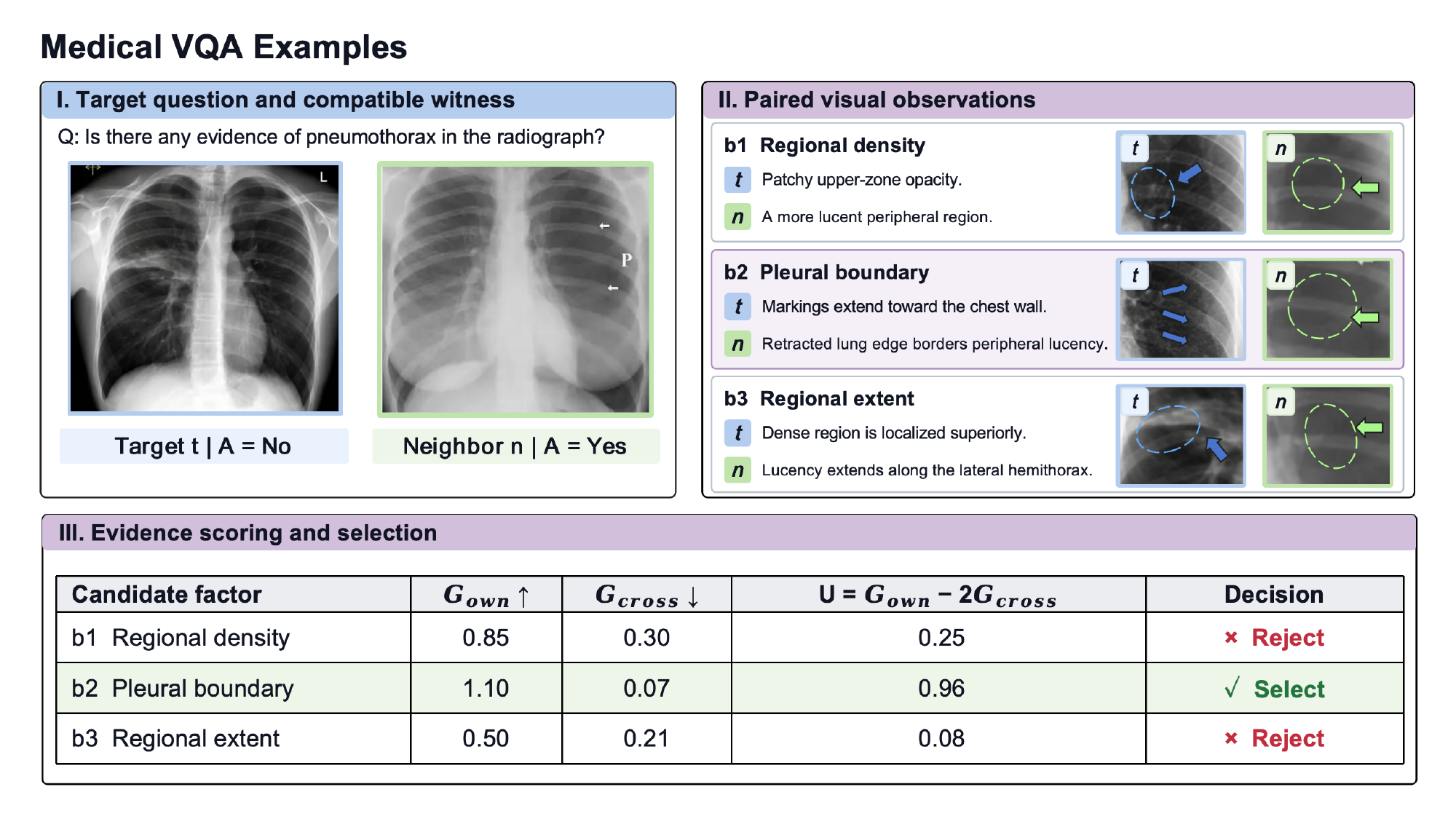}
\caption{Medical VQA example. For a target without pneumothorax and a
compatible witness with pneumothorax, BIRD selects the pleural-boundary
factor (b2), which achieves $G_{\mathrm{own}}=1.10$,
$G_{\mathrm{cross}}=0.07$, and $U=0.96$. Regional density and extent are
less boundary-specific and are therefore not selected.}
\label{fig:fig11}
\end{figure}

Figure~\ref{fig:fig11} presents a chest-radiograph question about
pneumothorax. The target is negative, while the retrieved witness is positive.
BIRD proposes paired observations concerning regional density, the pleural
boundary, and regional extent.

Regional density (b1) produces a relatively large own-side gain of $0.85$.
However, its substantial cross-boundary transfer of $0.30$ reduces its utility
to $0.25$, indicating that differences in opacity and lucency alone do not
reliably resolve the model's confusion. Regional extent (b3) is even less
effective, with a utility of $0.08$. The pleural-boundary factor (b2), by
contrast, identifies markings extending toward the chest wall in the negative
target and a retracted lung edge bordering peripheral lucency in the positive
witness. It yields the largest own-side gain
($G_{\mathrm{own}}=1.10$), limited cross-boundary transfer
($G_{\mathrm{cross}}=0.07$), and the highest utility ($U=0.96$).
BIRD consequently emphasizes the pleural-boundary distinction in the refined
rationale instead of relying on broader appearance differences that may occur
in both cases.

\subsection{Why High-Gain Evidence Can Still Be Bad}
\label{app:case_high_gain}

\begin{figure}[h]
\centering
\includegraphics[width=\linewidth]{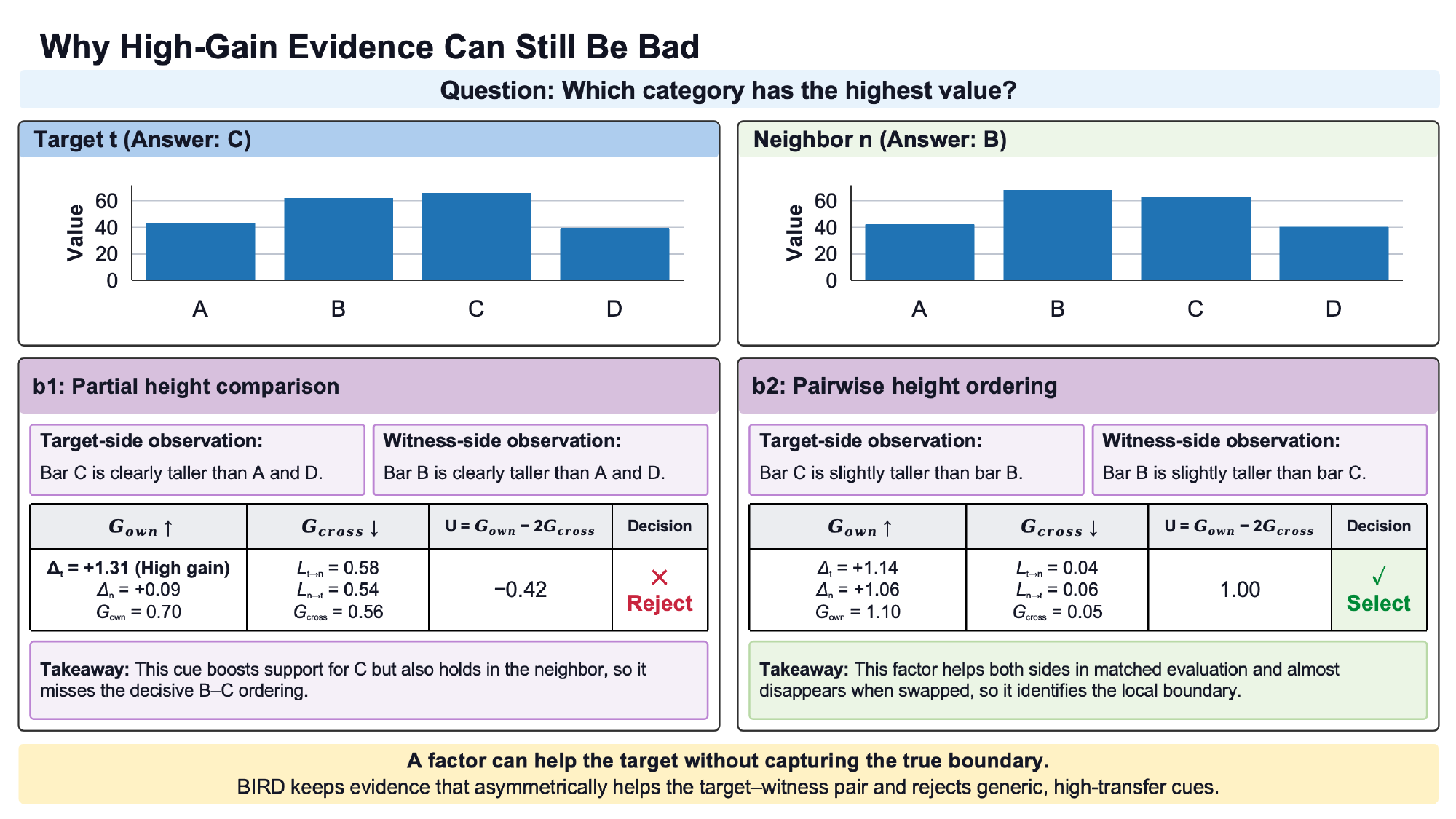}
\caption{Why high-gain evidence can still be unsuitable. The
partial-height factor (b1) strongly benefits the target but barely helps
the witness and transfers across the pair, resulting in negative boundary
utility. The pairwise B--C ordering (b2) supports the appropriate answer
on both sides and nearly disappears when swapped, allowing it to identify
the local decision boundary.}
\label{fig:fig12}
\end{figure}

Figure~\ref{fig:fig12} isolates why own-side gain alone is
insufficient for identifying boundary evidence. Both charts ask which
category has the highest value, but the target is answered C and the witness
is answered B. The partial-height factor (b1) appears useful when considering
only the target: stating that C is taller than A and D increases the target
model's correct-versus-confuser gap by $\Delta_t=1.31$. However, the
corresponding witness-side observation provides almost no support for the
witness's correct answer ($\Delta_n=0.09$). More importantly, these partial
comparisons remain applicable when transferred across the pair because both B
and C exceed A and D in both charts. The factor consequently has
$G_{\mathrm{cross}}=0.56$, and its utility falls from
$G_{\mathrm{own}}=0.70$ to $U=-0.42$.

Thus, this evidence raises confidence in the target answer without explaining
why the alternative answer is correct on the neighboring sample. It
strengthens a prediction but does not capture the boundary between B and C.
By comparison, the pairwise-ordering factor (b2) directly states that C is
slightly higher than B in the target and B is slightly higher than C in the
witness. It benefits both samples under matched evaluation
($\Delta_t=1.14$ and $\Delta_n=1.06$) but has little effect when swapped
($G_{\mathrm{cross}}=0.05$), producing $U=1.00$. This example illustrates why
BIRD jointly rewards own-side gain and penalizes cross-boundary transfer
instead of selecting evidence solely by its effect on the target.

\subsection{Model-Specific Example}
\label{app:case_model_specific}

\begin{figure}[h]
\centering
\includegraphics[width=\linewidth]{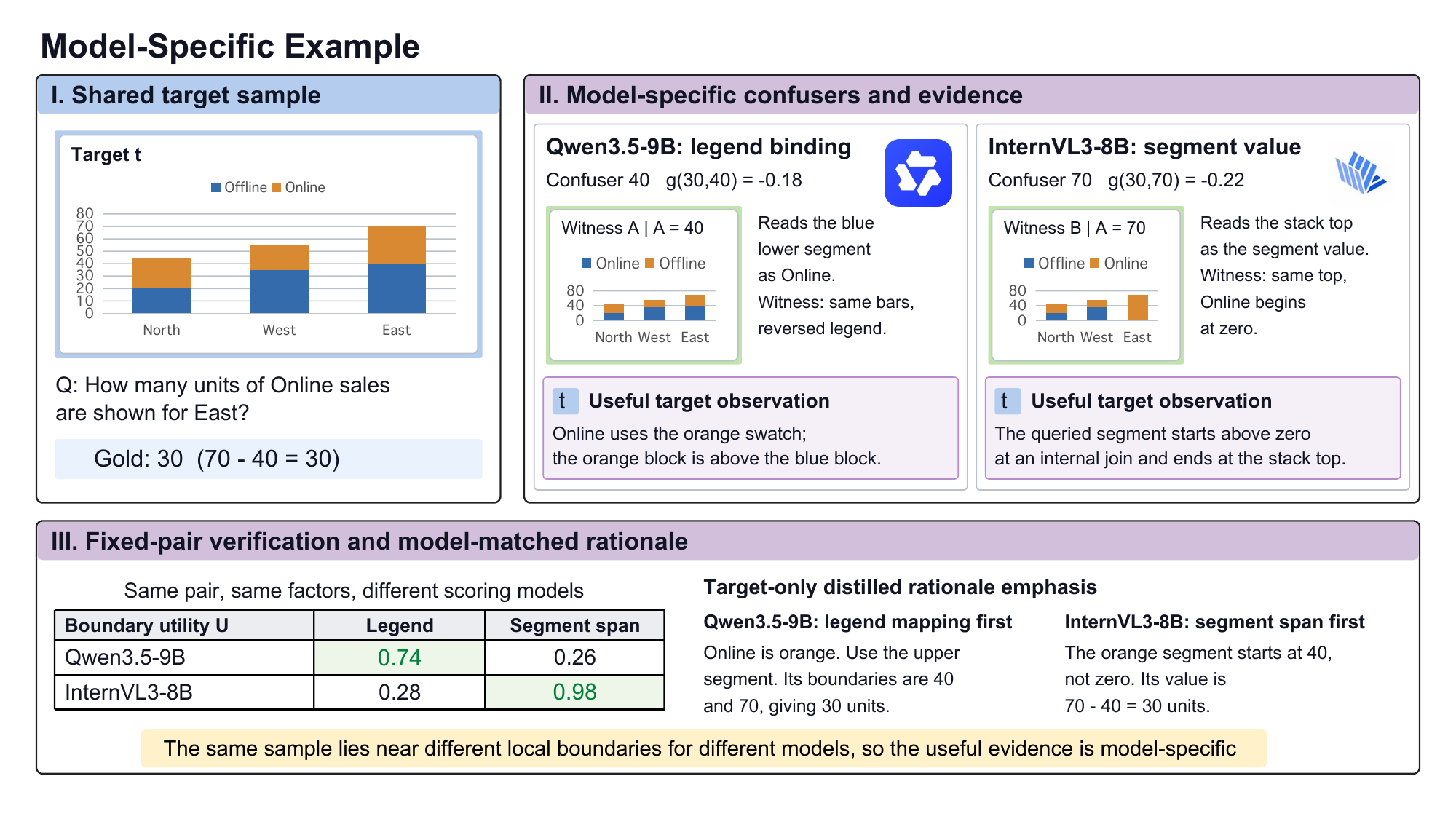}
\caption{Model-specific boundary evidence for a shared target sample.
Qwen3.5-9B confuses the gold answer $30$ with $40$ and assigns the highest
utility to legend-binding evidence, whereas InternVL3-8B confuses it with
$70$ and favors segment-span evidence. Even under fixed-pair verification,
the two models rank the same factors differently, leading to
model-matched rationale supervision.}
\label{fig:fig13}
\end{figure}

Figure~\ref{fig:fig13} demonstrates that the same sample can
expose different unresolved boundaries for different MLLMs. The shared target
asks for the Online sales value for East. Because the orange segment extends
from $40$ to $70$, the correct answer is $30$. Qwen3.5-9B instead prefers
$40$, with an energy gap of $g_\theta(30,40)=-0.18$, reflecting confusion
about the legend-to-segment mapping. Its retrieved witness preserves similar
bars but reverses the legend, and the useful target observation therefore
emphasizes that Online corresponds to the orange upper segment.

InternVL3-8B exhibits a different error on the same target. It prefers $70$,
with $g_\theta(30,70)=-0.22$, treating the top of the stack as the value of
the queried segment. Its witness has the same stack top but places the Online
segment from zero to $70$, exposing a confusion about segment span rather
than legend binding. The useful observation consequently emphasizes that the
target segment begins at the internal boundary of $40$ and ends at $70$.

The negative energy gaps show that each model initially favors its respective
confuser over the gold answer. Moreover, when the evaluated pair and candidate
factors are held fixed, the models still assign different boundary utilities.
Qwen3.5-9B assigns higher utility to legend evidence ($0.74$ versus $0.26$),
whereas InternVL3-8B favors segment-span evidence ($0.98$ versus $0.28$).
Their distilled rationales accordingly emphasize different reasoning steps:
legend mapping first for Qwen3.5-9B and segment subtraction first for
InternVL3-8B. The example shows that model specificity arises in both where
the unresolved boundary lies and which evidence is most effective for
resolving it.

\section{Limitations and Perspectives}
\label{sec:limitations_perspectives}

\subsection{Limitations}
\label{app:limitations}

BIRD has three main limitations. First, its boundary construction is local
and pairwise. For each target sample, BIRD selects one confusable alternative
and distills one verified factor. This design keeps the resulting supervision
focused and interpretable, but some decisions may involve several plausible
alternatives or require multiple complementary cues. A single
target--witness pair may not fully capture such cases.

Second, boundary utility provides a functional rather than causal test of
visual evidence. BIRD inserts a textual description of each candidate factor
and measures how it changes the target MLLM's answer preferences. This reveals
whether the factor helps distinguish the paired answers, but does not
establish that the model has visually grounded the corresponding feature.
The resulting scores may also be affected by answer phrasing, model
calibration, and the wording of the candidate factor. Boundary utility should
therefore be interpreted as a model-specific selection signal rather than
causal evidence of visual grounding.

Finally, our experiments cover two target MLLMs and two specialized domains:
chart/plot VQA and medical VQA. Although these settings differ substantially
in their visual content and reasoning requirements, further evaluation is
needed across additional model families, open-ended tasks, and specialized
domains. In particular, improvements on medical VQA benchmarks should not be
interpreted as establishing clinical reliability.

\subsection{Perspectives}
\label{app:perspectives}

BIRD currently constructs boundary-informed rationales in a single offline
round and then performs standard supervised fine-tuning. However, adaptation
may resolve some confusions while exposing others. A natural extension is
therefore to repeat neighbor retrieval, confuser selection, and evidence
verification after each adaptation stage. This would allow the supervision
to track the model's changing weaknesses while preserving single-sample
training and inference.

BIRD also examines each sample against one confusable alternative and
distills one factor. Future work could consider multiple competing
answers and combinations of complementary factors when a decision cannot be
resolved by a single visual distinction. The identified confusions could
further guide a model-specific data curriculum that prioritizes samples near
the model's remaining weaknesses. Evaluating these extensions across broader
model families, open-ended tasks, and additional specialized domains would
help establish the generality of boundary-informed supervision.

\end{document}